\documentclass[10pt,twocolumn,letterpaper]{article}
\usepackage{amsmath}
\usepackage{algorithm}
\usepackage{algpseudocode}
\usepackage{multirow}
\usepackage{rotating} 
\usepackage{booktabs}  
\usepackage{float}  
\usepackage[hypcap=false]{caption}
\PassOptionsToPackage{table,xcdraw,x11names}{xcolor}
\usepackage{cvpr} 
\usepackage[pagebackref,breaklinks,colorlinks,allcolors=cvprblue]{hyperref}
\definecolor{cvprblue}{rgb}{0.21,0.49,0.74}

\hypersetup{
    colorlinks=true,        
    linkcolor=customblue,   
    filecolor=customblue,   
    urlcolor=customblue,    
    citecolor=customblue    
}
\definecolor{tablecolor}{HTML}{ccf2f5} 
\definecolor{tablecolor2}{HTML}{ffcdb4}

\newcommand{\dd}[2]{$#1 \pm #2$} 
\newcommand{\ddbf}[2]{\textbf{$#1 \pm #2$}} 
\newcommand{\sdd}[2]{$#1 \pm #2$}

\definecolor{mycolor}{HTML}{BD8DE1}
\definecolor{customblue}{HTML}{D68594}

\title{{\huge Score and Distribution Matching Policy:}\\
\textmd{ Advanced Accelerated Visuomotor Policies via Matched Distillation}}

\author{
    Bofang Jia$^{1,2,*}$ \quad 
    Pengxiang Ding$^{1,3,*,\ddagger}$ \quad
    Can Cui$^{1,*}$ \quad
    Mingyang Sun$^{1,3}$ \\
    Pengfang Qian$^{1}$ \quad
    Siteng Huang$^{1}$ \quad
    Zhaoxin Fan$^{4}$ \quad
    Donglin Wang$^{1,\dagger}$ \\
    $^{1}$Westlake University \qquad
    $^{2}$Southwest University \qquad 
    $^{3}$Zhejiang University \qquad\\
    $^{4}$Beijing Advanced Innovation Center for Future Blockchain and Privacy Computing \\
    School of Artificial Intelligence, Beihang University \\
    {\tt \small jiabofang@email.swu.edu.cn, dingpx2015@gmail.com, cuican@westlake.edu.cn}\\
    {\tt\small \href{https://sdm-policy.github.io/}{\textbf{\textcolor{customblue}{Project Webpage}}}}
}

\begin{document}

\twocolumn[{%
 \renewcommand\twocolumn[1][]{#1}%
 \maketitle
 \vspace{-6mm}
 \centering
 \includegraphics[width=\textwidth]{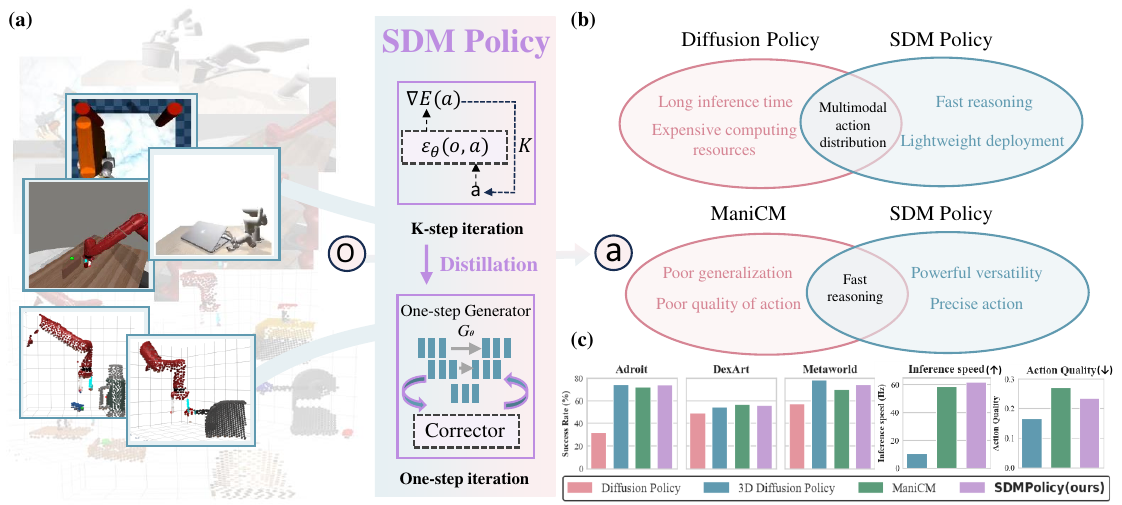}
 \vspace{-5mm}
 \captionof{figure}{ 
    {\bf SDM Policy}
    is a visual imitation learning algorithm that trains a one-step generator by enforcing a matching loss between two distributions. This approach balances fast inference speed and action accuracy, achieving state-of-the-art performance. (a) illustrates the principle of our method, (b) provides a comparison between SDM Policy, diffusion policy, and current SOTA methods (ManiCM), and (c) demonstrates that our method surpasses the current SOTA in task success rate and inference speed, showing that the quality of our actions is closer to the teacher model, resulting in more accurate action learning.%
    \label{fig:teaser}
   }
   \vspace{8mm}
}]

\maketitle

\renewcommand{\thefootnote}{\fnsymbol{footnote}}  
\footnotetext[1]{Equal contribution.}
\footnotetext[3]{Project Leader.}
\footnotetext[2]{Corresponding author.}

\begin{abstract}

Visual-motor policy learning has advanced with architectures like diffusion-based policies, known for modeling complex robotic trajectories. However, their prolonged inference times hinder high-frequency control tasks requiring real-time feedback. While consistency distillation (CD) accelerates inference, it introduces errors that compromise action quality. To address these limitations, we propose the Score and Distribution Matching Policy (SDM Policy), which transforms diffusion-based policies into single-step generators through a two-stage optimization process: score matching ensures alignment with true action distributions, and distribution matching minimizes KL divergence for consistency. A dual-teacher mechanism integrates a frozen teacher for stability and an unfrozen teacher for adversarial training, enhancing robustness and alignment with target distributions. Evaluated on a 57-task simulation benchmark, SDM Policy achieves a 6$\times$ inference speedup while having state-of-the-art action quality, providing an efficient and reliable framework for high-frequency robotic tasks. The code and more details can be found at 
\href{https://sdm-policy.github.io/}{\textbf{\textcolor{customblue}{sdm-policy.github.io}}}.

\end{abstract}

\section{Introduction}
\label{sec:intro}

Visual-motor policy learning has recently gained significant prominence in robotics, with a range of innovative architectures emerging for exploration, such as diffusion-based \cite{chi2023diffusion, ze20243d, lu2024manicm, ze2024generalizable}, flow-matching-based \cite{rouxel2024flow}, and transformer-based policies \cite{shafiullah2022behavior}. Diffusion-based policies are increasingly recognized for their capability to model intricate, high-dimensional robotic trajectories. This capability allows these policies to capture the complex temporal and spatial dependencies inherent in robotic tasks, providing a robust framework for generating diverse and feasible trajectories under varying task constraints. Therefore, diffusion-based policies have been widely applied to various tasks in robotics, such as grasping \cite{chi2023diffusion, ze20243d, ze2024generalizable}, and mobile manipulation \cite{ravan2024combining, yang2024equibot}.

However, diffusion-based policies inherently require prolonged inference times, as their step-by-step denoising process involves dozens or even hundreds of forward passes, consuming substantial computational resources and time. This poses a critical limitation for tasks demanding high-frequency control. For instance, tasks such as grasping objects, threading a needle, or picking up moving targets rely on real-time feedback to quickly adjust actions. In scenarios where objects slip or shift position, fast inference is crucial to make immediate corrections.

Currently, many efforts have been made to accelerate diffusion-based policies using consistency distillation (CD) \cite{lu2024manicm, prasad2024consistency}, which transfers knowledge from pre-trained teacher models into a single-step sampler. This approach enables rapid one-step generation and significantly improves inference speed.
However, CD-based methods are inherently prone to errors due to limitations in their underlying mechanisms. For instance, standard consistency distillation~\cite{song2023consistency} relies on ODE solvers~\cite{song2023consistency}, whose numerical approximations can introduce errors, resulting in suboptimal consistency and adversely affecting the model's overall performance. Similarly, latent consistency distillation~\cite{luo2023latent} is constrained by its focus on local consistency, which restricts the student model's ability to fully capture and integrate the comprehensive knowledge of the teacher model~\cite{kim2023consistency}.
These limitations lead to a loss of critical sample information, resulting in a decline in action quality and making it challenging to balance sample quality and inference speed. Consequently, achieving state-of-the-art inference speed while maintaining the accuracy and quality of sample actions is essential.

To solve these problems, we propose \textbf{Score and Distribution Matching Policy (SDM Policy)}, a framework designed to distill the capabilities of pre-trained diffusion-based teacher models into a single-step generator that is both efficient and accurate. The core of SDM Policy lies in its two-stage optimization process: score matching, which ensures that generated actions closely align with the true action distribution by leveraging the corrected score functions of diffusion policies, and distribution matching, which minimizes the KL divergence between the generator and the teacher policies to enforce distribution-level consistency. Unlike traditional iterative diffusion processes, SDM Policy enables the generator to directly produce high-quality actions in a single step, drastically reducing inference time. Additionally, the method incorporates a frozen teacher model as a stable reference and an unfrozen teacher model to guide the generator’s training through an adversarial optimization framework. This dual-teacher setup ensures the robustness of the generated actions while fostering alignment with the target distribution, thereby enhancing the overall reliability and generalizability of the model. The evaluation of SDM Policy on a 57-task simulation benchmark shows a 6× inference speedup with state-of-the-art action quality.

In summary, our contributions are three-fold:
\begin{enumerate}
    \item We introduce a framework, SDM Policy, that integrates score and distribution matching to transform diffusion-based policies into efficient single-step generators, enhancing inference speed while retaining action quality.
    \item We design a dual-teacher mechanism with a frozen teacher for stability and an unfrozen teacher for adversarial guidance, ensuring robustness and better alignment with the target action distribution.
    \item Extensive results shows the SDM Policy's effectiveness on a 57-task benchmark, achieving a 6× inference speedup over standard diffusion policies, with the state-of-the-art action quality.
\end{enumerate}

\section{Related Work}
\label{sec:related_work}

\subsection{Diffusion-based Robotic Policy}

Currently, visual-based policies~\cite{brohan2023rt1roboticstransformerrealworld, brohan2023rt2visionlanguageactionmodelstransfer, octomodelteam2024octoopensourcegeneralistrobot, wen2024tinyvlafastdataefficientvisionlanguageaction, kim2024openvlaopensourcevisionlanguageactionmodel, ding2025quar, song2024germ, liu2024rdt1bdiffusionfoundationmodel, ren2024diffusionpolicypolicyoptimization, chi2023diffusion, lu2024manicm, kim2023consistency, rouxel2024flow, zhang2024flowpolicyenablingfastrobust, zhao2023learningfinegrainedbimanualmanipulation, gong2024carp} have successfully tackled challenges in high-dimensional trajectory modeling and complex task decision-making. These advancements have led to the creation of robust and adaptable policy networks, enabling robotic systems to achieve substantial performance improvements across diverse tasks. Diffusion Policy~\cite{chi2023diffusion} is one of the pioneering works to explore this field, which represents robotic visual-motor policies as a conditional denoising process to generate actions, supporting high-dimensional action spaces and exhibiting impressive training stability. 3D Diffusion Policy~\cite{ze20243d} extends Diffusion Policy\cite{chi2023diffusion} to 3D scenarios by incorporating 3D visual information. It efficiently utilizes a compact 3D representation extracted from sparse point clouds with a high-performance point encoder, enhancing performance in robotic imitation learning. Additionally, HDP \cite{ma2024hierarchical}, DNAct \cite{yan2024dnact}, and IDP3 \cite{ze2024generalizable} extend Diffusion Policy \cite{chi2023diffusion} to more complex tasks, showcasing the powerful capability of diffusion policies in handling robotic tasks.

Despite the impressive performance of diffusion models, their costly inference speed has been a barrier for applications in robot tasks that require high real-time capabilities. To address this, a series of works focused on accelerating policies have been developed, and this work also falls into this category.

\subsection{Accelerated Robotic Policy}

Recently, many works have focused on accelerating diffusion policies \cite{lu2024manicm, prasad2024consistency, wang2024one, høeg2024streamingdiffusionpolicyfast}. The most classic approach to accelerating the policy is to use a consistency model~\cite{karras2022elucidating, liu2022pseudo, lu2022dpm, lu2022dpm++, zhao2024unipc, Duan_2023}, as it can directly map noise to data to generate high-quality samples, enabling fast one-step generation by design. Consistency policy~\cite{prasad2024consistency}, inspired by CTM~\cite{kim2023consistency}, denoising both back to the same time step $s$. Similarly, ManiCM~\cite{lu2024manicm} extends the one-step inference into 3D scenarios and achieves faster inference acceleration compared to 3D Diffusion Policy~\cite{ze20243d}. However, approaches to accelerate using consistency models often result in a loss of sample quality~\cite{yin2024improved}, making it challenging to balance speed and generation quality. Therefore, this paper is the first to explore an accelerated policy that maintains action quality while ensuring speed.

\section{Background}
\label{sec:Background}

\subsection{Task Formulation} Robotic manipulation is trained through imitation learning, with a small set of expert demonstrations containing complex skill trajectories utilized to learn a visuomotor policy $ \pi: \mathcal{O} \rightarrow \mathcal{A} $. This policy maps visual observations $ o \in \mathcal{O} $ to actions $ a \in \mathcal{A} $, enabling the robot to not only replicate expert skills but also generalize across different environments. 
The observation includes a combination of point clouds received from an eye-in-hand RGB-D camera and proprioceptive data from the robot or a combination of RGB camera images and proprioceptive data from the robot. The action space varies depending on the task and robot configuration, typically demonstrating SE(3) motion of the end-effector along with the standardized torque to be applied by the gripper fingers. 

\subsection{Diffusion Policy} Diffusion Policy \cite{chi2023diffusion} is an advanced vision-based motion policy for robots, designed to generate action sequences in complex tasks. This policy is achieved through a conditional denoising diffusion policy, where given conditions such as visual features and robot poses, random noise is gradually denoised into the target action sequence. Specifically, starting from a gaussian noise sample, the diffusion model utilizes a noise prediction network $\pi_\theta$ to predict and remove noise at each step, iterating for $T$ steps to generate a noise-free action $a^{0}$.
\small
\begin{equation}
a^{t-1} = \alpha_t \left( a^t - \gamma_t \pi_\theta(a^t, t, v, p) \right) + \sigma_t \mathcal{N}(0, I),
\end{equation}
\normalsize
where $a^t$ denotes the action at step $T$; $\alpha_t$, $\gamma_t$, and $\sigma_t$ are noise scheduling parameters controlling the denoising strength; $\mathcal{N}(0, I)$ is the noise prediction network used to estimate the noise at each step; $v$ and $p$ represent the visual features and robot poses as conditioning information; and $\mathcal{N}(0, I)$ is gaussian noise.

\section{Method}
\label{sec:method}

In this section, we first provide a detailed explanation of our pipeline and describe the design of each component in our SDM Policy (Section~\ref{sec:sdm}). To demonstrate the superiority of our approach, we analyze it in comparison with current methods and provide evidence of its effectiveness (Section~\ref{sec:discussion}).

\begin{figure*}[ht] 
    \centering
    \includegraphics[width=\textwidth]{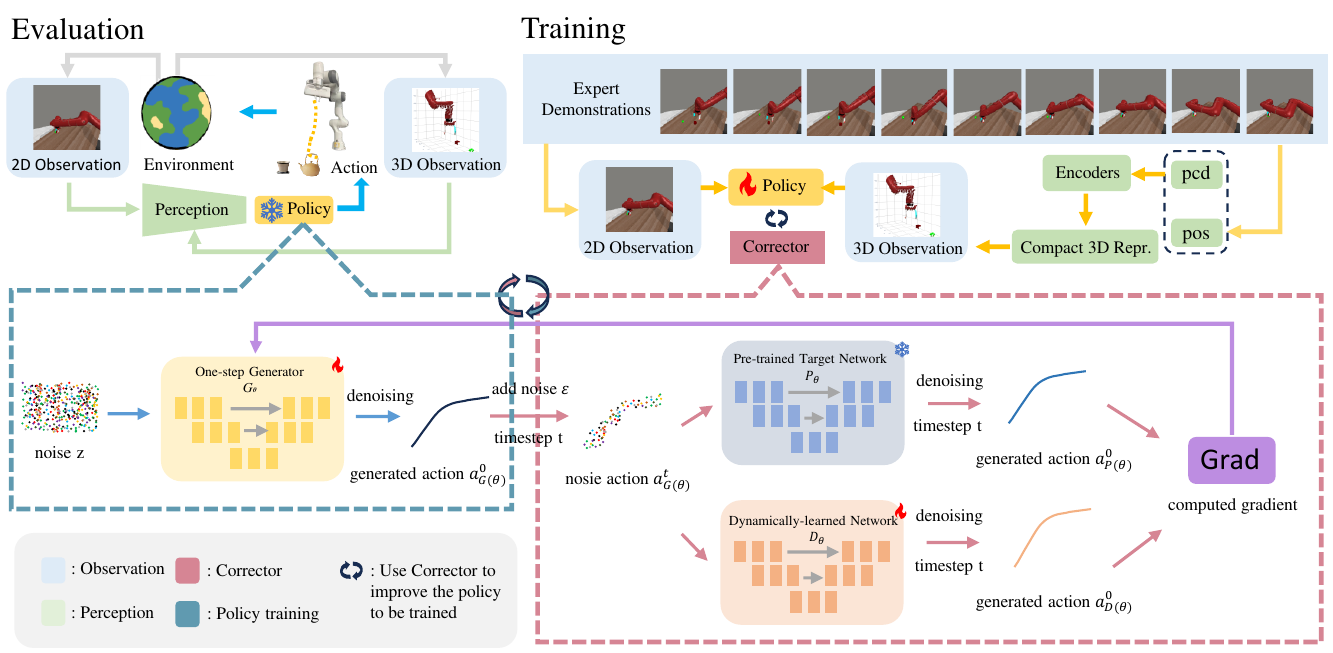} 
    \caption{\textbf{Overview of SDM Policy.} Our method distills diffusion policies, which require long inference times and high computational costs, into a fast and stable one-step generator. Our SDM Policy is represented by the one-step generator, which requires continual correction and optimization via the Corrector during training, but relies solely on the generator during evaluation. The corrector's optimization is based on two components: gradient optimization and diffusion optimization. The gradient optimization part primarily involves optimizing the entire distribution by minimizing the KL divergence between two distributions, $P_{\theta}$ and $D_{\theta}$, with distribution details represented through a score function that guides the gradient update direction, providing a clear signal. The diffusion optimization component enables $D_{\theta}$ to quickly track changes in the one-step generator’s output, maintaining consistency. Details on loading observational data for both evaluation and training processes are provided above the diagram. Our method applies to both 2D and 3D scenarios.}
    \label{fig:pipeline}
\end{figure*}

\subsection{Score and Distribution Matching Policy}
\label{sec:sdm}

The prolonged inference time of diffusion policies, due to the step-by-step denoising process, hinders their application in dynamic environments requiring high-frequency control and the practical deployment of lightweight robots. Accelerating the diffusion process to enable rapid action generation is essential. Addressing these challenges, our SDM policy achieves fast one-step generation through distribution matching, effectively resolving the slow inference issue of diffusion policies. We will provide a detailed explanation of the SDM policy's design and training optimization.

\subsubsection{Model Architecture}

Our SDM policy consists of two main components: the one-step generator and the corrector. The former is responsible for denoising pure noise input in a single step to restore precise actions, while the latter refines the one-step generator during training through gradient and diffusion optimization, ensuring it generates accurate actions comparable to those of the teacher model. The overall pipeline of SDM Policy is illustrated in Figure~\ref{fig:pipeline}, effectively addressing the low decision-making efficiency in diffusion policies.

\noindent\textbf{One-step generator.}
Our method distills the diffusion policies, which require long inference times and high computational costs, into a fast and stable one-step generator. The one-step generator starts from pure noise $z$ and generates accurate actions $a^0_\theta$ through single-step denoising, implemented by the generator $G_\theta$. To further improve the accuracy of action generation, we introduce a Corrector mechanism during training, which provides fine adjustments to the outputs of the one-step generator, ensuring the precision of the generated actions.

\noindent\textbf{Corrector.} To achieve more accurate action generation within the one-step generation framework, we introduce a unique Corrector structure. This Corrector consists of two networks, $P_{\theta}$ and $D_{\theta}$, which work together like adversarial generation. By comparing the outputs of these two networks, the Corrector determines the necessary action adjustments, serving as explicit labels for refinement. Both $P_{\theta}$ and $D_{\theta}$ are constructed based on the pre-trained diffusion policies $\pi_{\theta}$, with $P_{\theta}$ remaining fixed while $D_{\theta}$ continuously updates its parameters to adapt to the generator’s outputs.

To guide the learning process, we leverage the known properties of the diffusion model to approximate the score function over the diffusion distribution. This allows us to interpret the denoised output as the gradient direction, thereby guiding the Corrector’s adjustments. We use the KL divergence to measure the difference between the distributions represented by $P_{\theta}$ and $D_{\theta}$, providing detailed updates for the generator’s output action $a^0_{G(\theta)}$. This ensures that actions are generated in a more realistic direction. Finally, the gradient updates for our one-step generator are set to this difference, obtaining the necessary details for updating the generated actions and ensuring that the labels directly impact the training process of our one-step generator, gradually reducing the loss of learning information.

\small
\begin{multline}
D_{\text{KL}}\left( p_{D_{\theta}} \| p_{P_{\theta}} \right) 
= {\mathbb{E}} \left( \log \left( \frac{p_{D_{\theta}}(a_{G(\theta)}^{t})}{p_{P_{\theta}}(a_{G(\theta)}^{t})} \right) \right) \\
= \underset{z \sim \mathcal{N}(0, \mathbf{I})}{\mathbb{E}} \bigg[ - \big( \log p_{P_{\theta}}(a_{G(\theta)}^{t}) - \log p_{D_{\theta}}(a_{G(\theta)}^{t}) \big) \bigg].
\label{eq:2}
\end{multline}
\normalsize

\subsubsection{Training Stretrgy}

In the training phase, we progressively optimize the one-step generator through two components, enabling it to achieve impressive action generation.

\begin{figure}[ht] 
    \centering
    \includegraphics[width=0.48\textwidth]{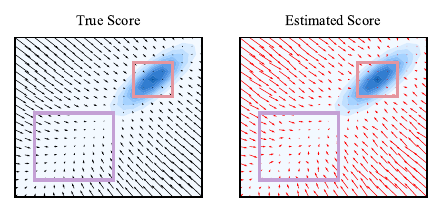} 
    \caption{\textbf{Performance of score estimation in low-density regions.} The purple rectangle represents low-density regions, and the pink rectangle represents high-density regions. For the entire rectangle, darker colors indicate higher density. The left image shows the true data scores, while the right image shows the estimated scores. In the high-density pink rectangle, the difference between the estimated and true scores is minimal. However, in the low-density purple rectangle, the difference between the estimated and true scores is significantly larger, indicating poor score matching performance in low-density regions. }
    \label{fig:low-di}
    \vspace{-0.8em}
\end{figure}

\noindent\textbf{Gradient optimization.} 
We face two challenges in gradient computation. First, real-world data often lies on a low-dimensional manifold within a high-dimensional ambient space \cite{song2019generative}. The score $\nabla_x \log p_{\text{data}}(x)$ represents the gradient in this ambient space, so it becomes undefined when $x$ is limited to the low-dimensional manifold. Consequently, data sparsity in low-density regions results in inaccurate score matching (Figure~\ref{fig:low-di}). Additionally, since our distributions use diffusion policy as a base model, the scores correspond to diffused distributions rather than the original, complicating accurate gradient estimation. To address these issues, we perturb the data with gaussian noise at varying levels, enabling score calculation in overlapping regions where both distributions' scores can be simultaneously computed \cite{song2019generative, yin2024onestep}.

Score computation refers to the process where, in the reverse denoising process, the model estimates the score $s(a^{t})$ to determine the denoising direction at each step, ultimately recovering the original noiseless sample. The score estimate $s(a^{t})$ provides a way to return from the noisy data $a^{t}$ to the noise-free action $a^{0}$. In practical applications \cite{song2021scorebased}, the score estimate can be expressed by the following formula:

\small
\begin{equation}
s(a^{t})=\frac{a^{t}-\alpha_{t} \pi_{\theta}(a^t, t, v, p)}{\sigma_{t}^{2}},
\label{eq:3}
\end{equation}
\normalsize
where \( \pi_{\theta} \) represents the trained diffusion policies,  $\alpha_t$ and $\sigma_t$ are noise scheduling parameters controlling the denoising strength, $v$ and $p$ represent the visual features and robot poses as conditioning information. Now, we only need the gradient with respect to $\theta$ to train our one-step generator through gradient descent.
\small
\begin{align}
\nabla_{\theta} D_{\text{KL}} &= \underset{z \sim \mathcal{N}(0, \mathbf{I})}{\mathbb{E}} \left[ - \left( s_{P_{\theta}}(a_{G(\theta)}^{t}) - s_{D_{\theta}}(a_{G(\theta)}^{t}) \right) \nabla_{\theta} G_{\theta}(z) \right] \notag \\
&= \underset{z \sim \mathcal{N}(0, \mathbf{I})}{\mathbb{E}} \left[ - \left( a_{P(\theta)}^{0} - a_{D(\theta)}^{0} \right) \nabla_{\theta} G_{\theta}(z) \right]
\label{eq:4}
\end{align}
\normalsize
\small
\begin{equation}
\mathcal{L}_{\text{one-step generator}} = \lambda \nabla_{\theta} D_{\text{KL}},
\label{eq:5}
\end{equation}
\normalsize
where $\lambda$ is a scaling factor for the one-step generator loss.

\noindent\textbf{Diffusion loss optimization.} For the dynamically changing $D_{\theta}$ in the corrector, the distribution of the generated actions changes throughout the training process. We need to continuously update $D_{\theta}$ to adapt to these changes, ensuring that the output $a_{D(\theta)}^{0}$ of $D_{\theta}$ remains consistent with the output $a_{G(\theta)}^{0}$ of the one-step generator. During training, we update the parameters $\theta$ by minimizing the standard denoising objective \cite{ho2020denoising, vincent2011connection}:
\small
\begin{equation}
\mathcal{L}_{\text{Diffusion}} = \gamma {\text{MSE}(a_{D(\theta)}^{0}, a_{G(\theta)}^{0})},
\label{eq:6}
\end{equation}
\normalsize
where $\gamma$ is a scaling factor for the diffusion loss.

\subsection{Foundations and Comparative Analysis}
\label{sec:discussion}

To better demonstrate the effectiveness of our method, we will conduct a discussion from two perspectives: theoretical analysis and comparative analysis with current state-of-the-art methods.

\noindent\textbf{Theoretical foundations.} To accelerate inference more effectively, numerous approaches have been attempted in the field of robotics for diffusion policies, including methods such as consistency distillation and latent consistency distillation for one-step generation. However, distillation for generative tasks often yields suboptimal results, failing to achieve the accurate action generation and multimodal action capabilities of the original diffusion policies. We found that this is because model optimization relies solely on distillation loss, lacking a direct signal similar to data labels in classification tasks. In this situation, the student model struggles to capture the details and diversity necessary for generating samples, resulting in decreased action accuracy and multimodal capability.

To address this shortcoming, our method offers an effective solution by leveraging classification to provide the one-step generator with direct supervision signals. Beyond simple distillation loss, we introduce a signal similar to classification loss to guide the policy toward realistic action generation, enabling continual optimization for greater accuracy. In our corrector structure, we use KL divergence to align the generated action distribution with the target distribution. This approach allows the one-step generator to learn an action policy consistent with the true distribution, even with limited expert demonstration data, thereby enhancing the diversity and generalization of generated behaviors.

\begin{figure}[!t]
    \centering
    \includegraphics[width=0.48\textwidth]{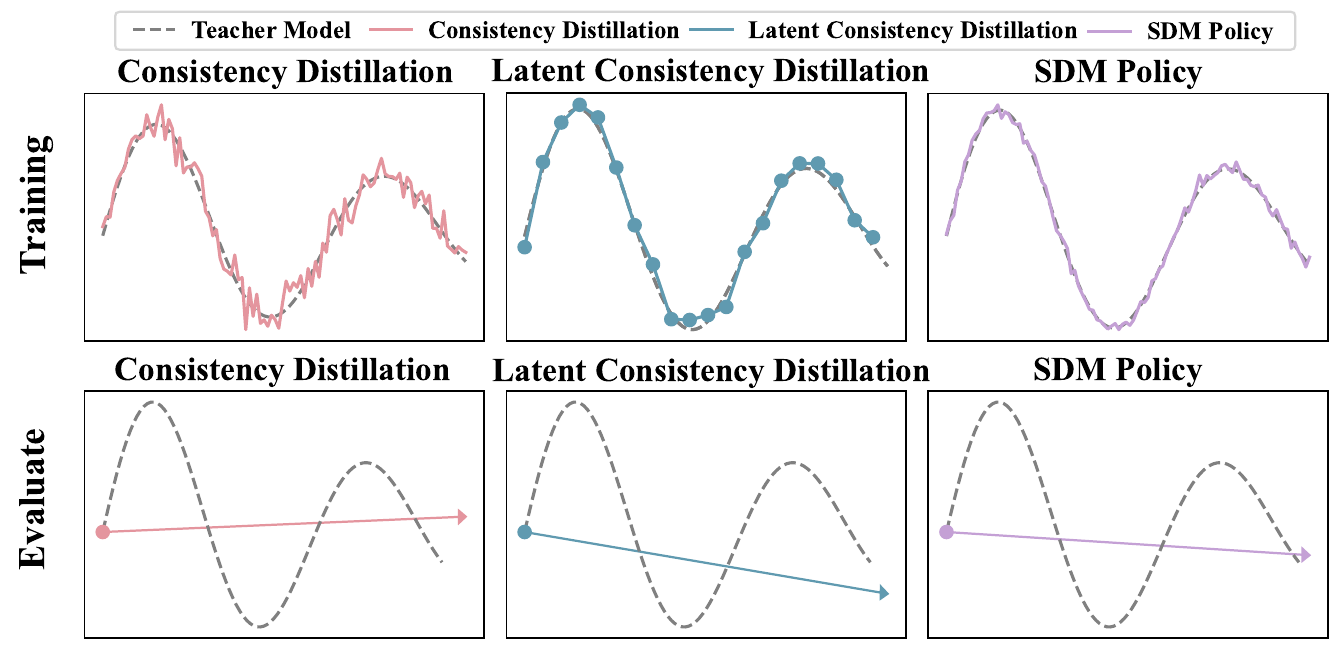} 
    \caption{\textbf{Comparison of SDM Policy and consistent distillation.} Here we provide a detailed comparison of the differences in the training process between consistency distillation, latent consistency distillation, and our SDM Policy. Consistency distillation suffers from significant deviations in one-step generation due to error accumulation, while latent consistency distillation quickly overlooks the need for global consistency. In contrast, our method aligns and learns at the distribution level, effectively addressing the issues mentioned above. }
    \label{fig:diffusion_policy_comparison}
    \vspace{-0.8em}
\end{figure}

\noindent\textbf{Limitations of existing methods.} The current state-of-the-art methods for accelerating diffusion policies are consistency distillation and latent consistency distillation, but both suffer from inevitable temporal errors, making it challenging to achieve accurate action generation and multimodal action generation capabilities through distillation in diffusion policies (Figure~\ref{fig:diffusion_policy_comparison}).

A major limitation of consistency distillation lies in the unavoidable accumulation of errors due to its reliance on a discrete consistency model. This discrete-time model is sensitive to the choice of $\Delta t$. The noise sample at the previous time step $t - \Delta t$, denoted as $a_{t-\Delta t}$, is derived from $a_t$ by solving the PF-ODE with a numerical ODE solver and a step size of $\Delta t$. This approach introduces discretization errors, resulting in inaccurate predictions during training. Since all solvers inherently introduce such errors, improving the solver is challenging, limiting the ability to achieve high-quality results in a few steps and hindering effective application.

In contrast, latent consistency distillation is constrained by local consistency, aligning only at intervals of $k$ steps. This introduces skip-step errors and aligns with the pre-trained model locally, lacking a global structural perspective. As a result, the quality and diversity of generated samples are compromised. Our method avoids these issues by eliminating stepwise error accumulation and local consistency limitations. Instead, we employ global consistency by learning the entire distribution of the diffusion policies and optimizing solely at the distribution level through KL divergence, achieving more efficient and robust improvement.

\section{Experiments}
\label{sec:experiments}

\begin{figure*}[ht]
    \centering
    \includegraphics[width=\textwidth]{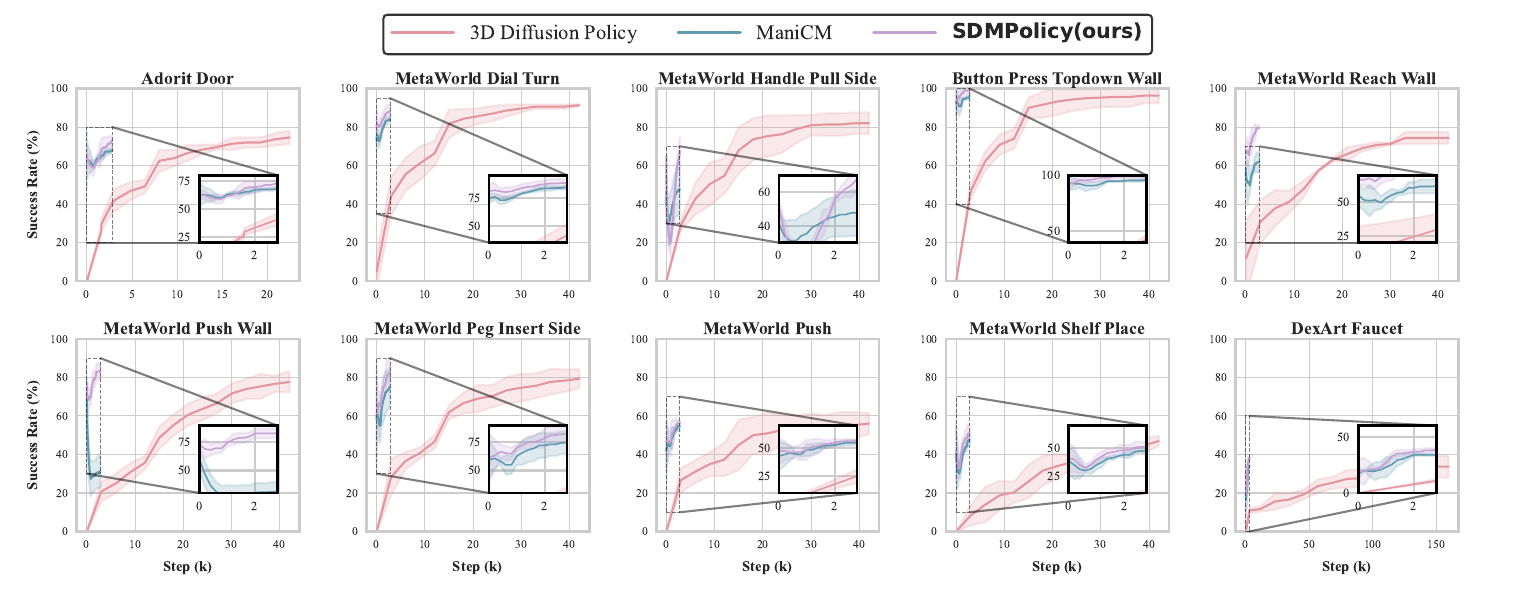} 
    \caption{\textbf{Learning efficiency.} We sampled 10 simulation tasks and presented the learning curves of our SDM Policy alongside DP3 and ManiCM. SDM Policy demonstrated a rapid convergence rate. In contrast, ManiCM showed slower learning progress, and DP3’s convergence speed was also slower than our method. }
    \label{fig:efficiency}
    \vspace{-1.em}
\end{figure*}

\begin{table*}[!t]
\centering
\caption{\small \textbf{Comparisons on success rate.} We evaluated $57$ challenging tasks using $3$ random seeds and reported the average success rate (\%) and standard deviation for three domains. $^{*}$ indicates our reproduction of that task. Our SDM Policy outperforms the current state-of-the-art model in one-step inference, achieving better results than Consistency Distillation and coming closer to the performance of the teacher model, which demonstrates its effectiveness.}
\label{table:mean_sr} 
\setlength{\tabcolsep}{4pt} 
\renewcommand{\arraystretch}{1.1} 
\small
\resizebox{\textwidth}{!}{
\begin{tabular}{l|c|cccccc|c}
\toprule
  Method & NFE & Adroit (3) & DexArt (4) & Metaworld Easy (28) & Metaworld Medium (11) & Metaworld Hard (6) & Metaworld Very Hard (5) & Average\\
\midrule
  Diffusion Policy~\cite{chi2023diffusion} & 10 & 31.7 & 49.0 & 83.6 & 31.1 & 9.0 & 26.6 & \sdd{55.5}{3.58} \\
  3D Diffusion Policy~\cite{ze20243d} & 10 & 68.3 & 68.5 & 90.9 & 61.6 & 31.7 & 49.0 & \sdd{72.6}{3.20} \\
  3D Diffusion Policy$^{*}$ & 10 & \textbf{74.3} & 54.3 & \textbf{89.0} & \textbf{72.7} & \textbf{38.0} & \textbf{75.8} & \sdd{\textbf{76.1}}{\textbf{2.32}} \\
  ManiCM$^{*}$ & 1 & 72.3 & \textbf{56.8} & 83.6 & 55.6 & 33.3 & 67.0 & \sdd{69.0}{4.60} \\
\midrule
\rowcolor{mycolor!10}
  \textbf{SDM Policy} & 1 & \underline{74.0} & \underline{56.0} & \underline{86.5} & \underline{65.8} & \underline{35.8} & \underline{71.6} & \underline{\sdd{74.8}{4.51}} \\
\bottomrule
\end{tabular}}
\end{table*}

\begin{table*}[ht]
\centering
\caption{\textbf{Success rate results for detailed experiments.} We provided a detailed presentation of the success rate (\%) and standard deviation for some tasks, with $^{*}$ indicating our reproduction of that task. The tasks were selected from different parts of the $57$ tasks, and we randomly chose $20$ tasks according to the corresponding proportion for demonstration. The results demonstrate the broad effectiveness of our approach on both simple grasping and pushing-pulling operations, as well as complex dexterous hand tasks.}
\label{table: 20tasks}

\resizebox{\textwidth}{!}{%
\begin{tabular}{cl|cc|cccccccc}
\toprule
& & \multicolumn{2}{c|}{Adroit} & \multicolumn{8}{c}{MetaWorld (Easy)}   \\
& Algorithm $\backslash$  Task & Door & Pen & Dial-Turn & Door-Unlock & Handle-Pull & Handle-Pull-Side & Lever-Pull & Reach-Wall & Window-Open & Peg-Unplug-Side  \\
\midrule

\multirow{5}{*}{\rotatebox{90}{3D Tasks}} 
& Diffusion Policy~\cite{chi2023diffusion} & \dd{37}{2} & \dd{13}{2} & \dd{63}{10} & \dd{98}{3} & \dd{27}{22} & \dd{23}{17} & \dd{49}{5} & \dd{59}{7} & \dd{100}{0} & \dd{74}{3}  \\

& 3D Diffusion Policy~\cite{ze20243d} & \dd{62}{4} & \dd{43}{6} & \dd{66}{1} & \dd{100}{0} & \dd{53}{11} & \dd{85}{3} & \dd{79}{8} & \dd{68}{3} & \dd{100}{0} & \dd{75}{5}  \\

& 3D Diffusion Policy$^{*}$ & \dd{\textbf{75}}{\textbf{3}} & \dd{\textbf{48}}{\textbf{3}} & \dd{\textbf{91}}{\textbf{0}} & \dd{\textbf{100}}{\textbf{0}} & \dd{\textbf{52}}{\textbf{8}} & \dd{\textbf{82}}{\textbf{5}} & \dd{\textbf{84}}{\textbf{8}} & \underline{\dd{74}{3}} & \dd{\textbf{99}}{\textbf{1}} & \dd{\textbf{93}}{\textbf{3}}  \\

& ManiCM$^{*}$ & \dd{68}{1} & \underline{\dd{49}{4}} & \dd{84}{2} & \dd{82}{16} & \dd{10}{10} & \dd{48}{11} & \dd{82}{7} & \dd{62}{5} & \underline{\dd{80}{26}} & \dd{71}{15}   \\

\rowcolor{mycolor!10}
& \textbf{SDM Policy (Ours)} & \underline{\dd{73}{2}} & \dd{42}{3} & \underline{\dd{88}{3}} & \underline{\dd{95}{6}} & \underline{\dd{28}{11}} & \underline{\dd{68}{6}} & \underline{\dd{84}{9}} & \dd{\textbf{80}}{\textbf{1}} & \dd{78}{18} & \underline{\dd{74}{19}}  \\

\bottomrule
\end{tabular}}

\resizebox{\textwidth}{!}{%
\begin{tabular}{cl|cccc|cc|cc|cc|cc}
\toprule
& & \multicolumn{4}{c|}{MetaWorld (Medium)} & \multicolumn{2}{c|}{MetaWorld (Hard)} & \multicolumn{2}{c|}{MetaWorld (Very Hard)} & \multicolumn{2}{c|}{DexArt} & \\
& Algorithm $\backslash$  Task & Peg-Insert-Side & Coffee-Pull & Push-Wall & Sweep & Pick-Out-Of-Hole & Push & Shelf-Place & Stick-Pull & Faucet & Bucket & Average \\
\midrule

\multirow{5}{*}{\rotatebox{90}{3D Tasks}} 
& Diffusion Policy~\cite{chi2023diffusion} & \dd{30}{5} & \dd{34}{7} & \dd{20}{3} & \dd{18}{8} & \dd{0}{0} & \dd{30}{3} & \dd{11}{3} & \dd{11}{2} & \dd{23}{8} & \dd{46}{1} & $38.4$ \\

& 3D Diffusion Policy~\cite{ze20243d} & \dd{42}{3} & \dd{87}{3} & \dd{49}{8} & \dd{96}{3} & \dd{14}{9} & \dd{51}{3} & \dd{17}{10} & \dd{27}{8} & \dd{63}{2} & \dd{46}{2} & $71.2$ \\

& 3D Diffusion Policy$^{*}$ & \underline{\dd{79}{4}} & \dd{\textbf{79}}{\textbf{2}} & \underline{\dd{78}{5}} & \dd{\textbf{92}}{\textbf{4}} & \dd{\textbf{44}}{\textbf{3}} & \underline{\dd{56}{5}} & \dd{47}{2} & \underline{\dd{67}{0}} & \dd{34}{5} & \dd{29}{2} & ${\textbf{70.1}}$ \\

& ManiCM$^{*}$ & \dd{75}{8} & \dd{68}{18} & \dd{31}{7} & \dd{54}{16} & \dd{30}{6} & \dd{55}{2} & \underline{\dd{48}{3}} & \dd{63}{2} & \underline{\dd{34}{0}} & \underline{\dd{36}{4}} & $56.7$ \\

\rowcolor{mycolor!10}
& \textbf{SDM Policy (Ours)} & \dd{\textbf{83}}{\textbf{5}} & \underline{\dd{72}{9}} & \dd{\textbf{83}}{\textbf{4}} & \underline{\dd{90}{6}} & \underline{\dd{34}{24}} & \dd{\textbf{57}}{\textbf{0}} & \dd{\textbf{51}}{\textbf{4}} & \dd{\textbf{68}}{\textbf{10}} & \dd{\textbf{38}}{\textbf{1}} & \dd{\textbf{31}}{\textbf{3}} & ${\underline{65.9}}$ \\

\bottomrule
\end{tabular}}

\end{table*}

\begin{table*}[!t]
\centering
\caption{\small \textbf{Comparisons on inference speed.} We evaluated $57$ challenging tasks using $3$ random seeds and reported the average speed (Hz) for three domains. $^{*}$ indicates our reproduction of that task, and - indicates that the data for this method has not been disclosed. Our SDM Policy outperforms the current state-of-the-art model in one-step inference, achieving better results than Consistency Distillation, providing strong evidence of the effectiveness of our model.}
\label{table:mean_time} 
\setlength{\tabcolsep}{4pt} 
\renewcommand{\arraystretch}{1.1} 
\small
\resizebox{\textwidth}{!}{
\begin{tabular}{l|c|cccccc|c}
\toprule
  Method & NFE & Adroit (3) & DexArt (4) & Metaworld Easy (28) & Metaworld Medium (11) & Metaworld Hard (6) & Metaworld Very Hard (5) & Average\\
\midrule
  Diffusion Policy~\cite{chi2023diffusion} & 10 & - & - & - & - & - & - &- \\
  3D Diffusion Policy~\cite{ze20243d} & 10 & - & - & - & - & - & - & - \\
  3D Diffusion Policy$^{*}$ & 10 & 10.79Hz & 10.62Hz & 10.01Hz & 10.79Hz & 10.92Hz & 10.53Hz & 10.39Hz \\
  ManiCM$^{*}$ & 1 & \textbf{57.56Hz} & \underline{73.19Hz} & \underline{55.03Hz} & \textbf{60.32Hz} & \textbf{66.26Hz} & \textbf{57.16Hz} & \underline{58.48Hz} \\
\midrule
\rowcolor{mycolor!10}
  \textbf{SDM Policy} & 1 & \underline{57.47Hz} & \textbf{75.59Hz} & \textbf{62.20Hz} & \underline{60.06Hz} & \underline{65.91Hz} & \underline{52.06Hz} & \textbf{61.75Hz} \\
\bottomrule
\end{tabular}}
\vspace{-1.6em}
\end{table*}

In this section, we first introduce the experimental setup, including the dataset, baseline methods, evaluation metrics, and implementation details(Section~\ref{exp:setup}). Then, we demonstrate the efficiency and effectiveness of our method in detail through experiments, along with visualized results from the experimental process(Section~\ref{exp:ee}). Finally, we conduct an ablation study to explore the impact of different design choices within our method(Section~\ref{exp:ablation}). In the main text, we provide a detailed description of the 57 tasks associated with 3D Diffusion Policy \cite{ze20243d}. For the 2D policy, we include it solely to demonstrate the generalizability of our method, indicating that it can distill any diffusion policies. Detailed results can be found in the supplementary material~\ref{sec: Implementation Details}.

\subsection{Experimental Setup}
\label{exp:setup}

To comprehensively and objectively evaluate our model, we set up 57 robotic tasks across three domains. This is primarily to ensure that our model is tested on a more scientifically rigorous benchmark, despite the fact that today's simulated tasks are increasingly realistic \cite{xiang2020sapien, zhu2020robosuite, todorov2012mujoco, ze20243d}.

\noindent \textbf{Datasets.} The 57 robotic tasks across three domains are sourced from Adroit \cite{rajeswaran2017learning}, DexArt \cite{bao2023dexart}, and MetaWorld \cite{yu2020meta}. Specifically, we use reinforcement learning VRL3 \cite{wang2022vrl3}, to obtain expert demonstrations for Adroit, while for MetaWorld tasks, we present results from scripted policies. According to \cite{seo2023masked}, MetaWorld tasks are categorized into different difficulty levels ranging from simple to very challenging. DexArt uses PPO \cite{schulman2017proximal} for trajectory generation. For each benchmark, we use a limited set of 10 expert demonstrations for training.

\noindent \textbf{Baselines.} To balance fast inference speed with accurate actions and to validate the effectiveness of our method, our benchmarks primarily include the advanced point cloud-based Diffusion Policy(DP3) \cite{ze20243d}, and one-step generation model ManiCM \cite{lu2024manicm}, generated through consistency distillation.

\noindent \textbf{Evaluation metrics.} For each random seed, we evaluate 20 segments every 200 training epochs, then calculate the average success rate of the top 5 segments as well as the average runtime for each task\footnote{All of our task training and evaluation were conducted on an NVIDIA A100 80G GPU.}. To ensure the fairness of the experiments, we randomly run three seeds for each task experiment, consistent with previous work \cite{lu2024manicm, ze20243d}.

\subsection{Efficiency and Effectiveness}
\label{exp:ee}

To demonstrate the effectiveness of our method, we conducted a comprehensive evaluation across four key aspects: task success rate, learning efficiency, inference time, and action learning. SDM Policy exhibited strong performance across all these areas. In the table, results highlighted in bold indicate the top-performing method, while results with an underline denote the second best performance across the various categories.

\noindent \textbf{High accuracy.} Table~\ref{table:mean_sr} presents the overall results for the 57 tasks across 3 domains. Compared to diffusion-based policies represented by DP3 and consistency-based policies represented by ManiCM, our method achieved a higher success rate across multiple tasks. Additionally, we randomly selected 20 tasks proportional to the number of tasks in each domain; the detailed results are shown in Table ~\ref{table: 20tasks}. A complete report of the success rates for each task can be found in the supplementary material.

\begin{figure*}[!t] 
    \centering
    \includegraphics[width=\textwidth]{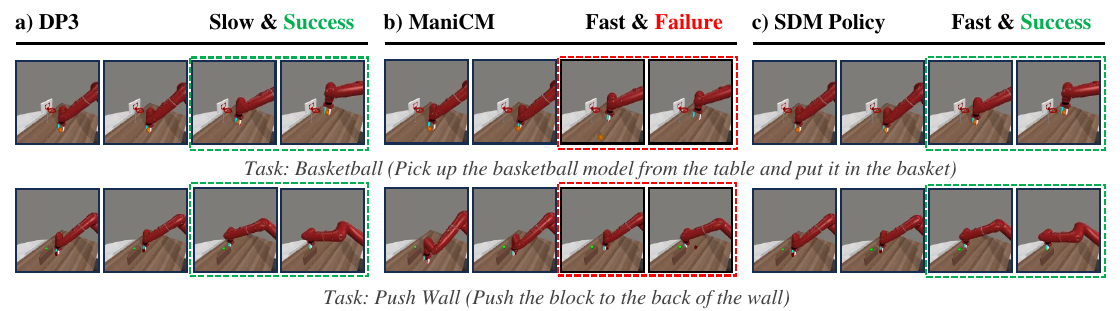} 
    \caption{\textbf{Task execution visualization.} Visualized keyframes for Basketball and Push Wall tasks, we found that during training, our SDM Policy could better complete the task and learn more precise actions, while consistency distillation might lead to task failure. }
    \label{fig:visualization}
    \vspace{-0.6em}
\end{figure*}

\noindent \textbf{Learning efficiency.} Figure~\ref{fig:efficiency} shows the results after 3000 epochs of training. We observed that the SDM Policy achieved a faster convergence rate compared to others. Furthermore, our SDM Policy distillation performance is closer to that of the teacher model than ManiCM.

\noindent \textbf{Competitive inference speed.} Table~\ref{table:mean_time} presents the overall inference speed results for the 57 tasks. Compared to diffusion-based policies represented by DP3 and consistency-based policies represented by ManiCM, our method achieves inference speeds on par with the state-of-the-art ManiCM, enabling rapid inference while being approximately 6 times faster than the diffusion-based DP3. Detailed inference times for individual tasks can be found in the supplementary material.

\begin{table}[!t]
\centering
\caption{\small \textbf{Comparisons on action.} We used $3$ random seeds to compare the learning effects of action generation on a $57$ challenging task during training, and reported the mean action error and standard deviation in the three domains. $^{*}$ indicates our reproduction of that task. Our SDM Policy outperforms current state-of-the-art models in one-step inference and achieves better results than Consistency Distillation.}
\label{table:mean_action} 
\setlength{\tabcolsep}{4pt} 
\renewcommand{\arraystretch}{1.1} 
\small
\resizebox{0.47\textwidth}{!}{
\begin{tabular}{l|c|ccc|c}
\toprule
  Method & NFE & Adroit (3) & DexArt (4) & Metaworld (50)  & Average\\
\midrule
  3D Diffusion Policy$^{*}$ & 10 &  \sdd{\textbf{0.055}}{\textbf{0.037}} & \sdd{0.017}{0.005}  & \sdd{\textbf{0.182}}{\textbf{0.311}} & \sdd{\textbf{0.166}}{\textbf{0.278}} \\
  ManiCM$^{*}$ & 1 & \underline{\sdd{0.179}{0.001}} & \sdd{\textbf{0.010}}{\textbf{0.000}}  & \sdd{0.298}{0.016} & \sdd{0.270}{0.014} \\
\midrule
\rowcolor{mycolor!10}
  \textbf{SDM Policy} & 1 & \sdd{0.179}{0.008}  & \sdd{\textbf{0.010}}{\textbf{0.000}} & \underline{\sdd{0.251}{0.019}} & \underline{\sdd{0.234}{0.017}} \\
\bottomrule
\end{tabular}}
\vspace{-0.6em}
\end{table}

\begin{table}[!t]
\centering
\caption{\small \textbf{Ablation on generator config.} We conduct training from scratch on the one-step generator to evaluate the impact of the pre-trained policy, reporting the overall task success rate, inference speed, and action quality. }
\vspace{-.4em}
\label{table:ablation} 
\setlength{\tabcolsep}{10pt} 
\renewcommand{\arraystretch}{1.1} 
\small
\resizebox{0.47\textwidth}{!}{
\begin{tabular}{l|ccc}
\toprule
  Method  & success rate  & inference speed & action  \\
\midrule
  SDM Policy  & \sdd{63.7}{2.7} & 65.44Hz  & \sdd{0.100}{0.007} \\
  w/o generator config  & \sdd{58.0}{4.0} & 51.19Hz  & \sdd{0.227}{0.013} \\
\bottomrule
\end{tabular}}
\vspace{-0.6em}
\end{table}

\noindent \textbf{Precise and accurate action.} As shown in Table~\ref{table:mean_action}, we also found that the action quality learned by SDM Policy is approximately better than that of ManiCM, the consistency-distilled model. This improvement indicates that our method preserves much more information during distillation, aligning with the previously noted higher success rates and learning efficiency, further validating the effectiveness of our approach. Detailed reports on action learning performance for individual tasks can be found in the supplementary material.

For task execution performance, we provide a detailed illustration in Figure~\ref{fig:visualization}. For certain tasks, our SDM Policy better captures the action execution of the teacher model, while the consistency distillation method may encounter task execution failures.

\subsection{Ablation Studies}
\label{exp:ablation}

In SDM Policy, we specifically load the parameters of the pre-trained policy $\pi_{\theta}$ to initialize our one-step generator $G_{\theta}$. To compare initialization methods for the one-step generator, validate, and explore our design choices, we selected seven tasks for experimentation: Adroit Door from Adroit, and Box-Close, Push, Reach, Reach Wall, Shelf-Place, and Sweep from MetaWorld. As shown in Table~\ref{table:ablation}, methods that do not load the parameters of the pre-trained policy $\pi_{\theta}$ fail to meet our requirements for rapid generation and high-quality actions, indicating the critical importance of properly utilizing pre-trained policy parameters.

\vspace{-0.2em}
\section{Discussion and Conclusion}
\label{sec:conclusion}

This work has demonstrated that enforcing the minimization of the matching loss between two diffusion distributions to provide direct signals for training the generator can significantly improve distillation performance. By addressing various robotic tasks in simulated environments, it has been shown that our method achieves a balance between fast inference speed and high-quality, precise actions, delivering SOTA results across all metrics. However, some unresolved issues remain worth exploring. All tasks in our simulation benchmark involve static objects and pre-defined environments, which limit the scope of validation to controlled settings. For practical applications, it is crucial to extend the approach to dynamic operations and tasks requiring high-frequency control to adapt to more complex and unpredictable environments. This may involve adopting different update frequencies to address challenges in our gradient optimization and diffusion optimization, which will serve as a primary direction for future work.

{
    \small
    \bibliographystyle{ieeenat_fullname}
    \bibliography{main}
}
\appendix
\renewcommand{\thesection}{\Alph{section}} 
\clearpage
\setcounter{page}{1}
\maketitlesupplementary

\section{Implementation Details}
\label{sec: Implementation Details}

\noindent\textbf{Task suite.} For the simulation experiments, to demonstrate the effectiveness of our method and ensure that our benchmarking is not influenced by the simulation environment, we conducted a comprehensive evaluation across 57 robotic tasks in three domains. These include 3 tasks from Adroit\cite{rajeswaran2017learning}, 4 tasks from DexArt\cite{bao2023dexart}, and 50 tasks from MetaWorld\cite{yu2020meta}, with the MetaWorld tasks categorized by varying difficulty levels. Table~\ref{table:task_suite} provides a brief overview, highlighting the differences in action dimensions, object morphology, and robot models among these tasks.

\begin{table}[htbp]
\caption{\textbf{Task suite.} Summarized 57 tasks in the simulation benchmark, including information on domains, robot models, object types, simulators, action dimensions, and the number of tasks. This demonstrates the diversity of the simulation benchmark in terms of robot types, object morphologies, simulators, and action dimensions, ensuring comprehensive evaluation across different scenarios.}
\label{table:task_suite}
\vspace{-0.05in}
\centering
\resizebox{0.48\textwidth}{!}{%
\begin{tabular}{lcccccc}
\toprule
Domain & Robot & Object & Simulator & Action Dimensions & Tasks Numbers \\
\midrule
\textbf{Adroit} & Shadow & Rigid/Articulated & MuJoCo & 28 & 3  \\
\textbf{DexArt} & Allegro & Articulated & Sapien & 22 & 4  \\
\textbf{MetaWorld} & Gripper & Rigid/Articulated & MuJoCo & 4 & 50  \\
\bottomrule
\end{tabular}
}
\vspace{-0.6em}
\end{table}

\noindent\textbf{Algorithm introduction.} Algorithm~\ref{alg:sdm_training} provides a detailed explanation of the training process for the SDM Policy, with a particular focus on the updated details of gradient optimization and diffusion optimization.

\begin{algorithm}[tbp]
\caption{SDM Policy Training Procedure}
\label{alg:sdm_training}
\textbf{Input:} One-step generator $G_\theta$, Pretrained Target Network $P_\theta$, Dynamically-learned Network $D_\theta$, and pre-trained diffusion policies $\pi_\theta$ \\
\textbf{Output:} Trained generator $G_\theta$. \\
\textbf{Initialization:} Initialize generator $G_\theta$, $P_\theta$, and $D_\theta$ from pretrained diffusion policies $\pi_\theta$ ($G_\theta \gets \pi_\theta$, $P_\theta \gets \pi_\theta$, $D_\theta \gets \pi_\theta$).
\begin{algorithmic}[1]
\While{not converged}
    \State // Generate action
    \State Sample $z \sim \mathcal{N}(0, I)$
    \State $a_{G(\theta)}^{0} \gets G_\theta(z)$
    \State // Add noise to action
    \State $a_{G(\theta)}^{t} \gets a_{G(\theta)}^{0}$

    \State // Compute the KL-divergence loss  (Eq.~\ref{eq:2})
    \State Compute the $ \mathcal{D}_{\text{KL}}$ between $D_\theta$ and $P_\theta$
    \State // Compute score function
    \State $s_{P_{\theta}}(a_{G(\theta)}^{t}) \gets \log p_{P_{\theta}}(a_{G(\theta)}^{t})$  (Eq.~\ref{eq:3})
    \State $s_{D_{\theta}}(a_{G(\theta)}^{t}) \gets \log p_{D_{\theta}}(a_{G(\theta)}^{t})$  (Eq.~\ref{eq:3})
    \State // Compute the one-step generator loss to update $G_\theta$ (Eq.~\ref{eq:5})
    \State Compute the $\mathcal{L}_{\text{one-step generator}}$ between $D_\theta$ and $P_\theta$
    \State // Compute the diffusion loss to update $D_\theta$ (Eq.~\ref{eq:6})
    \State Compute the $\mathcal{L}_{\text{diffusion}}$ between $D_\theta$ and $G_\theta$
    \If{iter mod $c == 0$}
    \State Update $G$, $D_\theta$
    \Else
    \State Update  $D_\theta$
    \EndIf

\EndWhile
\end{algorithmic}
\end{algorithm}

\section{All Results of Simulation Experiments}
\label{sec: Simulation Details}

To better demonstrate the effectiveness of SDM Policy, we provide a detailed presentation of the results related to task success rates, inference speed, and action performance discussed in the main text. The detailed results for individual tasks are separately reported in Table~\ref{table: simulation results with sr}, Table~\ref{table: simulation results with time}, Table~\ref{table: simulation results with action}. It is important to note that we did not calculate inference speed and action quality for tasks with a success rate of 0, as such data would be meaningless. In the subsequent tables, we uniformly represent these cases as ``Failure''.

\begin{table*}[htbp]
\centering
\caption{\textbf{Detailed results for 57 simulated tasks with success rates.} We evaluated 57 challenging tasks using 3 random seeds and reported the average success rate (\%) and standard deviation for each domain individually. $^{*}$ indicates our reproduction of the task. Our SDM policy outperforms the current state-of-the-art models in one-step inference, achieving better results than Consistency Distillation and coming closer to the performance of the teacher model, demonstrating its effectiveness.}
\label{table: simulation results with sr}

\resizebox{1.0\textwidth}{!}{%
\begin{tabular}{l|ccc|cccc|ccc}
\toprule
& \multicolumn{3}{c|}{\textbf{Adroit}} & \multicolumn{4}{c|}{\textbf{DexArt}} & \multicolumn{3}{c}{\textbf{Meta-World (Easy)}} \\

 Alg $\backslash$ Task & Hammer & Door & Pen & Laptop & Faucet & Toilet & Bucket & Button Press & Coffee Button  & Plate Slide Back Side\\
\midrule
Diffusion Policy  & \dd{45}{5} & \dd{37}{2} & \dd{13}{2} & \dd{69}{4} & \dd{23}{8} & \dd{58}{2} & \dd{46}{1} &\dd{99}{1} & \dd{99}{1} & \dd{100}{0}\\
3D Diffusion Policy & \ddbf{100}{0} & \ddbf{62}{4} & \ddbf{43}{6} & \ddbf{83}{1} & \ddbf{63}{2} & \ddbf{82}{4}  & \ddbf{46}{2} &\dd{100}{0} & \dd{100}{0} & \dd{100}{0} \\
3D Diffusion Policy$^{*}$ & \ddbf{100}{0} & \ddbf{75}{3} & \ddbf{48}{3} & \ddbf{80}{2} & \ddbf{34}{5} & \ddbf{74}{4}  & \ddbf{29}{2} & \dd{100}{0} & \dd{100}{0} & \dd{100}{0}\\
ManiCM$^{*}$ & \dd{100}{0} & \dd{68}{1} & \dd{49}{4} & \dd{83}{2} & \dd{34}{0} & \dd{74}{1} & \dd{36}{4} & \dd{100}{0} & \dd{100}{0} & \dd{100}{0}\\
SDM Policy  & \dd{100}{0} & \dd{73}{2} & \dd{49}{4} & \dd{83}{2} & \dd{38}{1} & \dd{72}{2} & \dd{31}{3} & \dd{100}{0} & \dd{100}{0} & \dd{100}{0}\\

\bottomrule
\end{tabular}}

\vspace{0.05cm}

\resizebox{1.0\textwidth}{!}{%
\begin{tabular}{l|ccccccc}
\toprule
& \multicolumn{6}{c}{\textbf{Meta-World (Easy)}} \\

 Alg $\backslash$ Task & Button Press Topdown & Button Press Topdown Wall & Button Press Wall & Peg Unplug Side  & Door Close & Door Lock\\
\midrule
Diffusion Policy  & \dd{98}{1} & \dd{96}{3} & \dd{97}{3} & \dd{74}{3} & \dd{100}{0} & \dd{86}{8} \\
3D Diffusion Policy  & \ddbf{100}{0} & \ddbf{99}{2} & \ddbf{99}{1} & \ddbf{75}{5} & \ddbf{100}{0} & \ddbf{98}{2} \\
3D Diffusion Policy$^{*}$  & \ddbf{99}{1} & \ddbf{96}{3} & \ddbf{100}{0} & \ddbf{93}{3} & \ddbf{100}{0} & \ddbf{96}{3} \\
ManiCM$^{*}$  & \dd{100}{0} & \dd{96}{2} & \dd{98}{3} & \dd{71}{15} & \dd{100}{0} & \dd{98}{2} \\
SDM Policy  & \dd{98}{2} & \dd{99}{1} & \dd{100}{0} & \dd{74}{19} & \dd{100}{0} & \dd{96}{2} \\

\bottomrule
\end{tabular}}

\vspace{0.05cm}

\resizebox{1.0\textwidth}{!}{%
\begin{tabular}{l|cccccccccc}
\toprule
& \multicolumn{8}{c}{\textbf{Meta-World (Easy)}} \\

 Alg $\backslash$ Task  & Door Open & Door Unlock & Drawer Close & Drawer Open & Faucet Close & Faucet Open & Handle Press & Handle Pull\\
\midrule
Diffusion Policy  & \dd{98}{3} & \dd{98}{3} & \dd{100}{0} & \dd{93}{3} & \dd{100}{0} & \dd{100}{0} & \dd{81}{4} & \dd{27}{22} \\
3D Diffusion Policy  & \ddbf{99}{1} & \ddbf{100}{0} & \ddbf{100}{0} & \ddbf{100}{0} & \ddbf{100}{0} & \ddbf{100}{0} & \ddbf{100}{0} & \ddbf{53}{11}\\
3D Diffusion Policy$^{*}$  & \ddbf{100}{0} & \ddbf{100}{0} & \ddbf{100}{0} & \ddbf{100}{0} & \ddbf{100}{0} & \ddbf{100}{0} & \ddbf{100}{0} & \ddbf{52}{8}\\
ManiCM$^{*}$  & \dd{100}{0} & \dd{82}{16} & \dd{100}{0} & \dd{100}{0} & \dd{100}{0} & \dd{100}{0} & \dd{100}{0} & \dd{10}{10}\\
SDM Policy  & \dd{100}{0} & \dd{100}{0} & \dd{100}{0} & \dd{100}{0} & \dd{99}{1} & \dd{100}{0} & \dd{100}{0} & \dd{28}{11}\\

\bottomrule
\end{tabular}}

\vspace{0.05cm}
\resizebox{1.0\textwidth}{!}{%
\begin{tabular}{l|ccccccccccc}
\toprule
& \multicolumn{8}{c}{\textbf{Meta-World (Easy)}} \\

 Alg $\backslash$ Task & Handle Press Side & Handle Pull Side & Lever Pull & Plate Slide & Plate Slide Back & Dial Turn  & Reach & Reach Wall\\
\midrule
Diffusion Policy  & \dd{100}{0} & \dd{23}{17} & \dd{49}{5} & \dd{83}{4} & \dd{99}{0} & \dd{63}{10} & \dd{18}{2} & \dd{59}{7} \\
3D Diffusion Policy  & \ddbf{100}{0} & \ddbf{85}{3} & \ddbf{79}{8} & \ddbf{100}{1} & \ddbf{99}{0} & \ddbf{66}{1} & \ddbf{24}{1} & \ddbf{68}{3}\\
3D Diffusion Policy$^{*}$  & \ddbf{0}{0} & \ddbf{82}{5} & \ddbf{84}{8} & \ddbf{100}{0} & \ddbf{100}{0} & \ddbf{91}{0} & \ddbf{26}{3} & \ddbf{74}{3}\\
ManiCM$^{*}$  & \dd{0}{0} & \dd{48}{11} & \dd{82}{7} & \dd{100}{0} & \dd{96}{5} & \dd{84}{2} & \dd{33}{3} & \dd{62}{5}\\
SDM Policy  & \dd{0}{0} & \dd{68}{6} & \dd{84}{9} & \dd{100}{0} & \dd{100}{0} & \dd{88}{3} & \dd{34}{3} & \dd{80}{1}\\

\bottomrule
\end{tabular}}

\vspace{0.05cm}

\resizebox{1.0\textwidth}{!}{%
\begin{tabular}{l|ccc|ccccc}
\toprule
& \multicolumn{3}{c|}{\textbf{Meta-World (Easy)}} & \multicolumn{5}{c}{\textbf{Meta-World (Medium)}} \\

 Alg $\backslash$ Task & Plate Slide Side & Window Close & Window Open  & Basketball & Bin Picking & Box Close & Coffee Pull & Coffee Push\\
\midrule
Diffusion Policy  & \dd{100}{0} & \dd{100}{0} & \dd{100}{0} & \dd{85}{6} & \dd{15}{4} & \dd{30}{5} & \dd{34}{7} & \dd{67}{4}\\
3D Diffusion Policy  & \ddbf{100}{0} & \ddbf{100}{0} & \ddbf{100}{0} & \ddbf{98}{2} & \dd{34}{30} & \ddbf{42}{3} & \ddbf{87}{3} & \ddbf{94}{3}\\

3D Diffusion Policy$^{*}$  & \ddbf{100}{0} & \ddbf{100}{0} & \ddbf{99}{1} & \ddbf{100}{0} & \dd{56}{14} & \ddbf{59}{5} & \ddbf{79}{2} & \ddbf{96}{2}\\
ManiCM$^{*}$  & \dd{100}{0} & \dd{100}{0} & \dd{80}{26} & \dd{4}{4} & \dd{49}{17} & \dd{73}{2} & \dd{68}{18} & \dd{96}{3}\\
SDM Policy   & \dd{100}{0} & \dd{100}{0} & \dd{78}{18} & \dd{28}{26} & \dd{55}{13} & \dd{61}{3} & \dd{72}{9} & \dd{97}{2}\\

\bottomrule
\end{tabular}}

\vspace{0.05cm}

\resizebox{1.0\textwidth}{!}{%
\begin{tabular}{l|cccccc|ccc}
\toprule
& \multicolumn{6}{c|}{\textbf{Meta-World (Medium)}} & \multicolumn{3}{c}{\textbf{Meta-World (Hard)}} \\

 Alg $\backslash$ Task & Hammer & Peg Insert Side & Push Wall & Soccer & Sweep & Sweep Into & Assembly & Hand Insert & Pick Out of Hole \\
\midrule
Diffusion Policy  & \dd{15}{6} & \dd{34}{7} & \dd{20}{3} & \dd{14}{4} & \dd{18}{8} & \dd{10}{4} & \dd{15}{1} & \dd{0}{0} & \dd{0}{0} \\
3D Diffusion Policy & \ddbf{76}{4} & \ddbf{69}{7} & \ddbf{49}{8} & \ddbf{18}{3} & \ddbf{96}{3} & \ddbf{15}{5} & \ddbf{99}{1} & \ddbf{14}{4} & \ddbf{14}{9} \\
3D Diffusion Policy$^{*}$ & \ddbf{100}{0} & \ddbf{79}{4} & \ddbf{78}{5} & \ddbf{23}{4} & \ddbf{92}{4} & \ddbf{38}{9} & \ddbf{100}{0} & \ddbf{28}{8} & \ddbf{44}{3} \\
ManiCM$^{*}$  & \dd{98}{2} & \dd{75}{8} & \dd{31}{7} & \dd{27}{3} & \dd{54}{16} & \dd{37}{13} & \dd{87}{3} & \dd{28}{15} & \dd{30}{16} \\

SDM Policy  & \dd{98}{2} & \dd{83}{5} & \dd{83}{4} & \dd{25}{2} & \dd{90}{6} & \dd{32}{15} & \dd{100}{0} & \dd{24}{14} & \dd{34}{24} \\

\bottomrule
\end{tabular}}

\vspace{0.05cm}

\resizebox{1.0\textwidth}{!}{%
\begin{tabular}{l|ccc|ccccc|ccccc}
\toprule
& \multicolumn{3}{c|}{\textbf{Meta-World (Hard)}} & \multicolumn{5}{c|}{\textbf{Meta-World (Very Hard)}} & \multicolumn{5}{c}{\textbf{Average}} \\

 Alg $\backslash$ Task & Pick Place & Push & Push Back & Shelf Place & Disassemble & Stick Pull & Stick Push & Pick Place Wall & \\
\midrule
Diffusion Policy & \dd{0}{0}& \dd{30}{3} & \dd{0}{0} & \dd{11}{3} & \dd{43}{7} & \dd{11}{2} & \dd{63}{3} & \dd{5}{1} & \dd{55.5}{3.58}\\
3D Diffusion Policy  & \ddbf{12}{4} & \ddbf{51}{3} & \ddbf{0}{0} & \dd{17}{10} & \ddbf{69}{4} & \ddbf{27}{8} & \ddbf{97}{4} & \ddbf{35}{8}  & \ddbf{72.6}{3.20}\\
3D Diffusion Policy$^{*}$  & \ddbf{0}{0} & \ddbf{56}{5} & \ddbf{0}{0} & \dd{47}{2} & \ddbf{91}{4} & \ddbf{67}{0} & \ddbf{100}{0} & \ddbf{74}{4} & \ddbf{76.1}{2.32}\\
ManiCM$^{*}$ & \dd{0}{0} & \dd{55}{2} & \dd{0}{0} & \dd{48}{3} & \dd{87}{3} & \dd{63}{2} & \dd{100}{0} & \dd{37}{16} & \ddbf{69.0}{4.60}\\
SDM Policy & \dd{0}{0} & \dd{57}{0} & \dd{100}{0} & \dd{51}{4} & \dd{86}{10} & \dd{68}{10} & \dd{0}{0} & \dd{53}{12} & \dd{74.8}{4.51}\\

\bottomrule
\end{tabular}}

\end{table*}

\begin{table*}[!t]
\centering
\caption{\textbf{Detailed results for 57 simulated tasks with inference speed.} We evaluated 57 challenging tasks using 3 random seeds and reported the average inference speed (Hz) for each domain individually. $^{*}$ indicates our reproduction of that task, and to ensure fairness we must use the same computing resource configuration. Our SDM policy outperforms the current state-of-the-art models in one-step inference, achieving better results than Consistency Distillation, demonstrating its effectiveness.}
\label{table: simulation results with time}

\resizebox{1.0\textwidth}{!}{%
\begin{tabular}{l|ccc|cccc|ccc}
\toprule
& \multicolumn{3}{c|}{\textbf{Adroit}} & \multicolumn{4}{c|}{\textbf{DexArt}} & \multicolumn{3}{c}{\textbf{Meta-World (Easy)}} \\

 Alg $\backslash$ Task & Hammer & Door & Pen & Laptop & Faucet & Toilet & Bucket & Button Press & Coffee Button  & Plate Slide Back Side\\
\midrule

3D Diffusion Policy$^{*}$ & 10.78Hz & 11.40Hz & 10.18Hz & 10.19Hz & 10.58Hz & 10.72Hz  & 9.99Hz & 10.20Hz & 10.25Hz & 10.31Hz\\
ManiCM$^{*}$ & 74.36Hz & 48.80Hz & 49.51Hz & 72.50Hz & 78.19Hz & 70.95Hz & 71.12Hz & 50.69Hz & 53.56Hz & 66.35Hz\\
SDM Policy  & 76.56Hz & 48.80Hz & 47.04Hz & 70.23Hz & 82.85Hz & 75.20Hz & 74.08Hz & 49.96Hz & 59.47Hz & 83.08Hz\\

\bottomrule
\end{tabular}}

\vspace{0.05cm}

\resizebox{1.0\textwidth}{!}{%
\begin{tabular}{l|ccccccc}
\toprule
& \multicolumn{6}{c}{\textbf{Meta-World (Easy)}} \\

 Alg $\backslash$ Task & Button Press Topdown & Button Press Topdown Wall & Button Press Wall & Peg Unplug Side  & Door Close & Door Lock\\
\midrule

3D Diffusion Policy$^{*}$  & 10.28Hz & 10.27Hz & 10.40Hz & 10.12Hz & 10.35Hz & 10.14Hz \\
ManiCM$^{*}$  & 52.49Hz & 53.16Hz & 53.00Hz & 55.64Hz & 51.78Hz & 53.97Hz \\
SDM Policy  & 52.51Hz & 51.85Hz & 52.56Hz & 66.78Hz & 51.42Hz & 54.15Hz \\

\bottomrule
\end{tabular}}

\vspace{0.05cm}

\resizebox{1.0\textwidth}{!}{%
\begin{tabular}{l|cccccccccc}
\toprule
& \multicolumn{8}{c}{\textbf{Meta-World (Easy)}} \\

 Alg $\backslash$ Task  & Door Open & Door Unlock & Drawer Close & Drawer Open & Faucet Close & Faucet Open & Handle Press & Handle Pull\\
\midrule

3D Diffusion Policy$^{*}$  & 10.12Hz & 10.16Hz & 10.65Hz & 10.19Hz & 10.39Hz & 10.28Hz & 10.31Hz & 10.20Hz\\
ManiCM$^{*}$  & 50.93Hz & 66.25Hz & 51.97Hz & 54.03Hz & 53.73Hz & 50.98Hz & 51.82Hz & 51.59Hz\\
SDM Policy  & 50.51Hz & 52.09Hz & 52.23Hz & 50.89Hz & 56.25Hz & 51.92Hz & 52.13Hz & 89.75Hz\\

\bottomrule
\end{tabular}}

\vspace{0.05cm}
\resizebox{1.0\textwidth}{!}{%
\begin{tabular}{l|ccccccccccc}
\toprule
& \multicolumn{8}{c}{\textbf{Meta-World (Easy)}} \\

 Alg $\backslash$ Task & Handle Press Side & Handle Pull Side & Lever Pull & Plate Slide & Plate Slide Back & Dial Turn  & Reach & Reach Wall\\
\midrule

3D Diffusion Policy$^{*}$  & Failure & 10.25Hz & 10.10Hz & 10.27Hz & 10.12Hz & 10.27Hz & 10.04Hz & 10.11Hz\\
ManiCM$^{*}$  & Failure & 51.73Hz & 51.76Hz & 50.78Hz & 51.21Hz & 83.92Hz & 55.92Hz & 51.80Hz\\
SDM Policy  & Failure & 53.35Hz & 55.16Hz & 81.32Hz & 83.71Hz & 87.74Hz & 88.08Hz & 64.95Hz\\

\bottomrule
\end{tabular}}

\vspace{0.05cm}

\resizebox{1.0\textwidth}{!}{%
\begin{tabular}{l|ccc|ccccc}
\toprule
& \multicolumn{3}{c|}{\textbf{Meta-World (Easy)}} & \multicolumn{5}{c}{\textbf{Meta-World (Medium)}} \\

 Alg $\backslash$ Task & Plate Slide Side & Window Close & Window Open  & Basketball & Bin Picking & Box Close & Coffee Pull & Coffee Push\\
\midrule

3D Diffusion Policy$^{*}$  & 10.33Hz & 10.55Hz & 10.50Hz & 10.64Hz & 11.42Hz & 10.57Hz & 11.54Hz & 11.73Hz\\
ManiCM$^{*}$  & 50.56Hz & 63.4Hz & 52.88Hz & 56.07Hz & 70.41Hz & 64.61Hz & 64.56Hz & 51.86Hz\\
SDM Policy   & 81.42Hz & 50.50Hz & 55.60Hz & 52.04Hz & 79.31Hz & 64.41Hz & 64.80Hz & 91.63Hz\\

\bottomrule
\end{tabular}}

\vspace{0.05cm}

\resizebox{1.0\textwidth}{!}{%
\begin{tabular}{l|cccccc|ccc}
\toprule
& \multicolumn{6}{c|}{\textbf{Meta-World (Medium)}} & \multicolumn{3}{c}{\textbf{Meta-World (Hard)}} \\

 Alg $\backslash$ Task & Hammer & Peg Insert Side & Push Wall & Soccer & Sweep & Sweep Into & Assembly & Hand Insert & Pick Out of Hole \\
\midrule

3D Diffusion Policy$^{*}$ & 10.38Hz & 10.31Hz & 10.55Hz & 10.57Hz & 10.48Hz & 10.46Hz & 10.53Hz & 10.80Hz & 10.60Hz \\
ManiCM$^{*}$  & 49.94Hz & 68.25Hz & 61.42Hz & 54.44Hz & 54.16Hz & 65.1Hz & 62.67Hz & 69.48Hz & 66.51Hz \\

SDM Policy  & 49.32Hz & 57.06Hz & 50.38Hz & 47.82Hz & 52.07Hz & 51.78Hz & 65.92Hz & 53.36Hz & 58.91Hz \\

\bottomrule
\end{tabular}}

\vspace{0.05cm}

\resizebox{1.0\textwidth}{!}{%
\begin{tabular}{l|ccc|ccccc|ccccc}
\toprule
& \multicolumn{3}{c|}{\textbf{Meta-World (Hard)}} & \multicolumn{5}{c|}{\textbf{Meta-World (Very Hard)}} & \multicolumn{5}{c}{\textbf{Average}} \\

 Alg $\backslash$ Task & Pick Place & Push & Push Back & Shelf Place & Disassemble & Stick Pull & Stick Push & Pick Place Wall & \\
\midrule

3D Diffusion Policy$^{*}$  & Failure & 10.75Hz & Failure & 10.56Hz & 10.46Hz & 10.40Hz & 10.54Hz & 10.69Hz & 10.39Hz\\

ManiCM$^{*}$ & Failure & 65.80Hz & Failure & 72.32Hz & 54.18Hz & 54.92Hz & 51.18Hz & 53.18Hz & 58.48Hz\\
SDM Policy & Failure & 85.47Hz & Failure & 54.27Hz & 51.06Hz & 52.90Hz & 49.57Hz & 52.52Hz & 61.75Hz\\

\bottomrule
\end{tabular}}
\vspace{1.0em}
\end{table*}
\begin{table*}[!t]
\centering
\caption{\textbf{Detailed results for 57 simulated tasks with action.} We used $3$ random seeds to compare the learning effects of action generation on a $57$ challenging task during training, and reported the mean action error and standard deviation in the three domains. $^{*}$ indicates our reproduction of that task, and to ensure fairness we must use the same computing resource configuration. Our SDM policy outperforms the current state-of-the-art models in one-step inference, achieving better results than Consistency Distillation, demonstrating its effectiveness.}
\label{table: simulation results with action}

\resizebox{1.0\textwidth}{!}{%
\begin{tabular}{l|ccc|cccc|ccc}
\toprule
& \multicolumn{3}{c|}{\textbf{Adroit}} & \multicolumn{4}{c|}{\textbf{DexArt}} & \multicolumn{3}{c}{\textbf{Meta-World (Easy)}} \\

 Alg $\backslash$ Task & Hammer & Door & Pen & Laptop & Faucet & Toilet & Bucket & Button Press & Coffee Button  & Plate Slide Back Side\\
\midrule

3D Diffusion Policy$^{*}$ & \ddbf{0.041}{0.025} & \ddbf{0.054}{0.029} & \ddbf{0.055}{0.038} & \ddbf{0.020}{0.007} & \ddbf{0.016}{0.005} & \ddbf{0.019}{0.005}  & \ddbf{0.014}{0.004} & \dd{0.077}{0.053} & \dd{0.034}{0.007} & \dd{0.025}{0.008}\\

ManiCM$^{*}$ & \dd{0.033}{0.000} & \dd{0.016}{6.803} & \dd{0.490}{0.001} & \dd{0.010}{7.370} & \dd{0.009}{3.659} & \dd{0.010}{9.961} & \dd{0.010}{4.237} & \dd{0.171}{0.009} & \dd{0.066}{0.001} & \dd{0.011}{0.000}\\
SDM Policy  & \dd{0.032}{9.204} & \dd{0.006}{2.703} & \dd{0.500}{0.000} & \dd{0.002}{0.103} & \dd{0.017}{0.001} & \dd{0.007}{0.001} & \dd{0.012}{0.000} & \dd{0.141}{0.003} & \dd{0.069}{0.001} & \dd{0.007}{4.247}\\

\bottomrule
\end{tabular}}

\vspace{0.05cm}

\resizebox{1.0\textwidth}{!}{%
\begin{tabular}{l|ccccccc}
\toprule
& \multicolumn{6}{c}{\textbf{Meta-World (Easy)}} \\

 Alg $\backslash$ Task & Button Press Topdown & Button Press Topdown Wall & Button Press Wall & Peg Unplug Side  & Door Close & Door Lock\\
\midrule

3D Diffusion Policy$^{*}$  & \ddbf{0.102}{0.115} & \ddbf{0.067}{0.076} & \ddbf{0.110}{0.040} & \ddbf{0.121}{0.063} & \ddbf{0.109}{0.276} & \ddbf{0.257}{0.243} \\
ManiCM$^{*}$  & \dd{0.158}{0.002} & \dd{0.166}{0.001} & \dd{0.173}{0.002} & \dd{0.195}{0.006} & \dd{0.776}{0.021} & \dd{0.186}{0.007} \\
SDM Policy  & \dd{0.114}{0.001} & \dd{0.153}{0.001} & \dd{0.140}{0.001} & \dd{0.200}{0.004} & \dd{0.056}{0.000} & \dd{0.0143}{0.003} \\

\bottomrule
\end{tabular}}

\vspace{0.05cm}

\resizebox{1.0\textwidth}{!}{%
\begin{tabular}{l|cccccccccc}
\toprule
& \multicolumn{8}{c}{\textbf{Meta-World (Easy)}} \\

 Alg $\backslash$ Task  & Door Open & Door Unlock & Drawer Close & Drawer Open & Faucet Close & Faucet Open & Handle Press & Handle Pull\\
\midrule

3D Diffusion Policy$^{*}$  & \ddbf{0.081}{0.070} & \ddbf{0.193}{0.168} & \ddbf{0.124}{0.138} & \ddbf{0.057}{0.012} & \ddbf{0.151}{0.126} & \ddbf{0.149}{0.124} & \ddbf{0.394}{0.539} & \ddbf{0.612}{1.355}\\
ManiCM$^{*}$  & \dd{0.092}{0.001} & \dd{0.339}{0.011} & \dd{0.676}{0.011} & \dd{0.095}{0.002} & \dd{0.295}{0.007} & \dd{0.308}{0.006} & \dd{0.884}{0.098} & \dd{2.888}{0.142}\\
SDM Policy  & \dd{0.089}{0.001} & \dd{0.296}{0.005} & \dd{0.626}{0.017} & \dd{0.100}{0.001} & \dd{0.295}{0.002} & \dd{0.248}{0.002} & \dd{0.877}{0.052} & \dd{2.622}{0.205}\\

\bottomrule
\end{tabular}}

\vspace{0.05cm}
\resizebox{1.0\textwidth}{!}{%
\begin{tabular}{l|ccccccccccc}
\toprule
& \multicolumn{8}{c}{\textbf{Meta-World (Easy)}} \\

 Alg $\backslash$ Task & Handle Press Side & Handle Pull Side & Lever Pull & Plate Slide & Plate Slide Back & Dial Turn  & Reach & Reach Wall\\
\midrule

3D Diffusion Policy$^{*}$  & Failure & \ddbf{1.355}{8.273} & \ddbf{0.156}{0.069} & \ddbf{0.096}{0.021} & \ddbf{0.032}{0.009} & \ddbf{0.050}{0.009} & \ddbf{0.006}{0.001} & \ddbf{0.080}{0.029}\\
ManiCM$^{*}$  & Failure & \dd{2.414}{0.236} & \dd{0.042}{0.001} & \dd{0.213}{0.002} & \dd{0.046}{0.001} & \dd{0.055}{0.000} & \dd{0.041}{4.470} & \dd{0.045}{0.001}\\
SDM Policy  & Failure & \dd{2.298}{0.490} & \dd{0.005}{8.085} & \dd{0.163}{0.001} & \dd{0.056}{0.000} & \dd{0.040}{0.004} & \dd{0.094}{0.014} & \dd{0.019}{0.000}\\

\bottomrule
\end{tabular}}

\vspace{0.05cm}

\resizebox{1.0\textwidth}{!}{%
\begin{tabular}{l|ccc|ccccc}
\toprule
& \multicolumn{3}{c|}{\textbf{Meta-World (Easy)}} & \multicolumn{5}{c}{\textbf{Meta-World (Medium)}} \\

 Alg $\backslash$ Task & Plate Slide Side & Window Close & Window Open  & Basketball & Bin Picking & Box Close & Coffee Pull & Coffee Push\\
\midrule

3D Diffusion Policy$^{*}$  & \ddbf{0.137}{0.028} & \ddbf{0.179}{0.208} & \ddbf{0.351}{0.250} & \ddbf{0.293}{0.276} & \dd{0.183}{0.094} & \ddbf{0.168}{0.092} & \ddbf{0.045}{0.014} & \ddbf{0.077}{0.017}\\
ManiCM$^{*}$  & \dd{0.143}{0.007} & \dd{0.230}{0.017} & \dd{0.485}{0.030} & \dd{0.539}{0.034} & \dd{0.057}{0.002} & \dd{0.037}{0.001} & \dd{0.029}{0.000} & \dd{0.050}{0.001}\\
SDM Policy   & \dd{0.157}{0.002} & \dd{0.202}{0.003} & \dd{0.521}{0.013} & \dd{0.357}{0.004} & \dd{0.027}{0.000} & \dd{0.102}{0.022} & \dd{0.017}{5.617} & \dd{0.049}{0.001}\\

\bottomrule
\end{tabular}}

\vspace{0.05cm}

\resizebox{1.0\textwidth}{!}{%
\begin{tabular}{l|cccccc|ccc}
\toprule
& \multicolumn{6}{c|}{\textbf{Meta-World (Medium)}} & \multicolumn{3}{c}{\textbf{Meta-World (Hard)}} \\

 Alg $\backslash$ Task & Hammer & Peg Insert Side & Push Wall & Soccer & Sweep & Sweep Into & Assembly & Hand Insert & Pick Out of Hole \\
\midrule

3D Diffusion Policy$^{*}$ & \ddbf{0.032}{0.010} & \ddbf{0.327}{0.274} & \ddbf{0.237}{0.224} & \ddbf{0.035}{0.010} & \ddbf{0.288}{0.236} & \ddbf{0.261}{0.106} & \ddbf{0.040}{0.011} & \ddbf{0.087}{0.010} & \ddbf{0.340}{0.174} \\
ManiCM$^{*}$  & \dd{0.051}{0.001} & \dd{0.101}{0.007} & \dd{0.063}{0.001} & \dd{0.024}{5.471} & \dd{0.117}{0.010} & \dd{0.059}{0.003} & \dd{0.233}{0.018} & \dd{0.044}{0.000} & \dd{0.413}{0.008} \\

SDM Policy  & \dd{0.055}{0.000} & \dd{0.089}{0.002} & \dd{0.005}{2.332} & \dd{0.004}{5.203} & \dd{0.174}{0.007} & \dd{0.038}{0.000} & \dd{0.059}{0.000} & \dd{0.036}{0.000} & \dd{0.267}{0.003} \\

\bottomrule
\end{tabular}}

\vspace{0.05cm}

\resizebox{1.0\textwidth}{!}{%
\begin{tabular}{l|ccc|ccccc|ccccc}
\toprule
& \multicolumn{3}{c|}{\textbf{Meta-World (Hard)}} & \multicolumn{5}{c|}{\textbf{Meta-World (Very Hard)}} & \multicolumn{5}{c}{\textbf{Average}} \\

 Alg $\backslash$ Task & Pick Place & Push & Push Back & Shelf Place & Disassemble & Stick Pull & Stick Push & Pick Place Wall & \\
\midrule

3D Diffusion Policy$^{*}$  & Failure & \ddbf{0.040}{0.008} & Failure & \dd{0.416}{0.377} & \ddbf{0.107}{0.025} & \ddbf{0.218}{0.120} & \ddbf{0.088}{0.020} & \ddbf{0.212}{0.217} & \ddbf{0.166}{0.278}\\
ManiCM$^{*}$ & Failure & \dd{0.056}{0.001} & Failure & \dd{0.197}{0.011} & \dd{0.089}{0.001} & \dd{0.089}{0.002} & \dd{0.123}{0.003} & \dd{0.397}{0.039} & \ddbf{0.270}{0.014}\\
SDM Policy & Failure & \dd{0.044}{0.000} & Failure & \dd{0.261}{0.005} & \dd{0.076}{0.000} & \dd{0.057}{0.000} & \dd{0.102}{0.001} & \dd{0.229}{0.012} & \dd{0.234}{0.017}\\

\bottomrule
\end{tabular}}
\vspace{1.0em}
\end{table*}

\begin{figure*}[!t] 
    \centering
    \includegraphics[width=\textwidth]{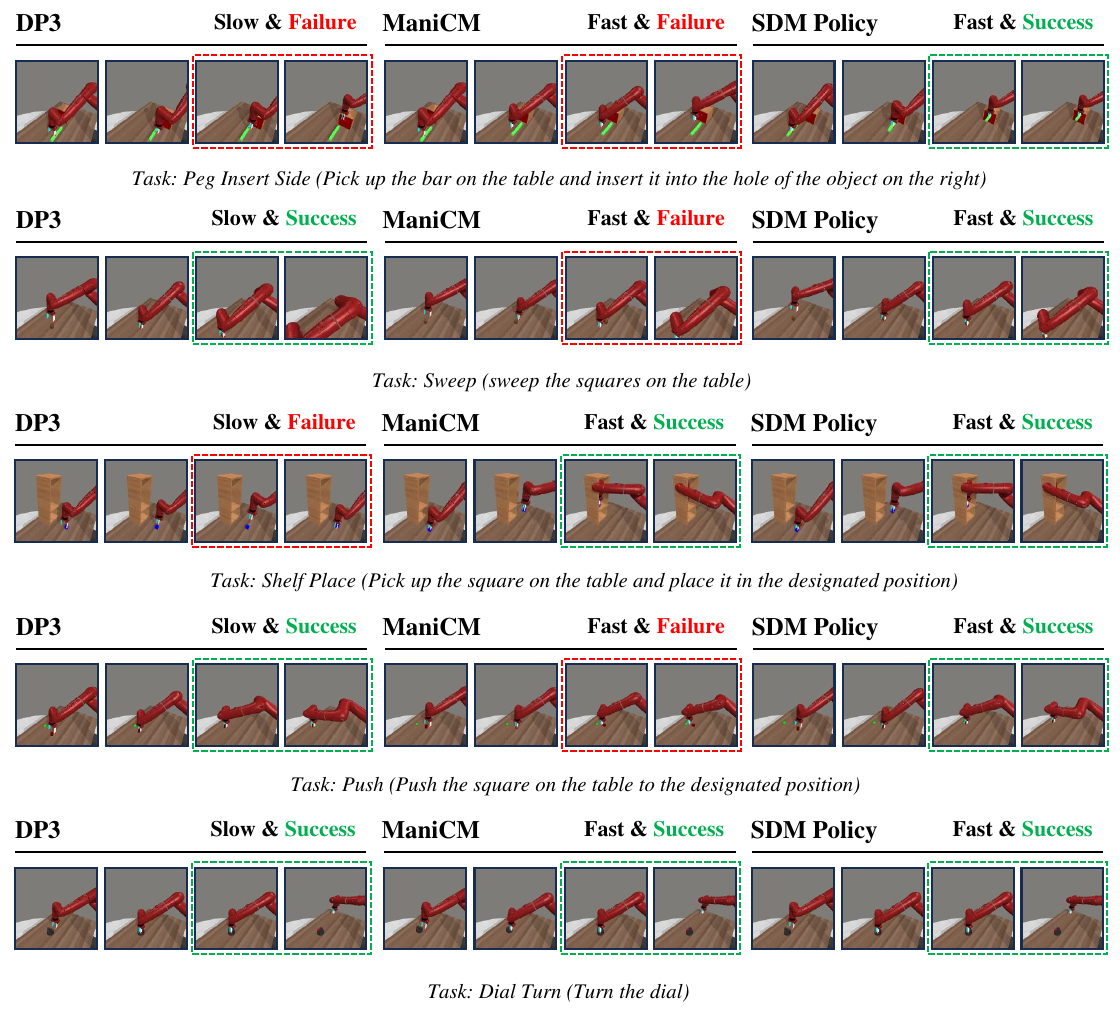} 
    \caption{\textbf{Part 1 of the task execution visualization.} We conducted a detailed visualization of the performance on a subset of 57 tasks. Specifically, 10 tasks were randomly selected for demonstration and comparison based on the domain and task difficulty, proportionally representing the 57 tasks. This is the first part of the task performance visualization, focusing on tasks categorized as medium, hard, and very hard in MetaWorld. The demonstrations are taken from an intermediate state across all epochs, with red indicating task failure and green indicating task success. The dashed boxes highlight keyframes of either failures or successes. For most tasks, our SDM Policy can complete the tasks quickly and accurately, demonstrating the effectiveness of our method. }
    \label{fig:visualization_more_1}
    \vspace{-0.6em}
\end{figure*}

\begin{figure*}[!t] 
    \centering
    \includegraphics[width=\textwidth]{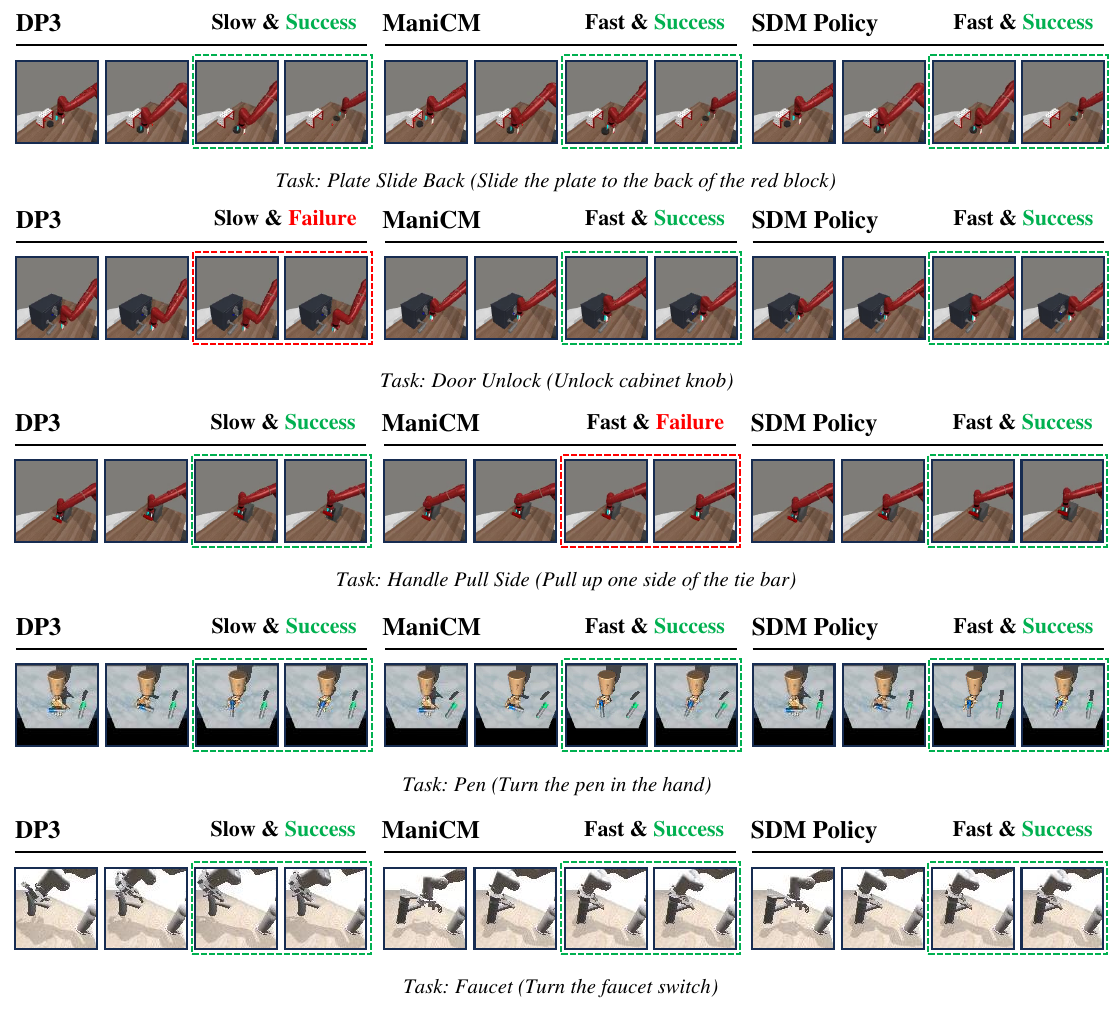} 
    \caption{\textbf{Part 2 of the task execution visualization.} We conducted a detailed visualization of the performance on a subset of 57 tasks. Specifically, 10 tasks were randomly selected for demonstration and comparison based on the domain and task difficulty, proportionally representing the 57 tasks. This is the second part of the task performance visualization, focusing on tasks categorized as easy in MetaWorld, Adroit and DexArt. The demonstrations are taken from an intermediate state across all epochs, with red indicating task failure and green indicating task success. The dashed boxes highlight keyframes of either failures or successes. For most tasks, our SDM Policy can complete the tasks quickly and accurately, demonstrating the effectiveness of our method.}
    \label{fig:visualization_more_2}
    \vspace{-0.6em}
\end{figure*}

For task execution performance, we provide a more detailed illustration in Figure~\ref{fig:visualization_more_1} and Figure~\ref{fig:visualization_more_2}. For certain tasks, our SDM Policy better captures the action execution of the teacher model, while the consistency distillation method may encounter task execution failures. For the demonstration of these tasks, we adopt a unified intermediate state of epochs, which indirectly proves the effectiveness of our method in terms of learning efficiency mentioned in Section ~\ref{exp:ee} in the main text.

\section{SDM Policy in 2D Scene}
\label{sec:universality}

To better demonstrate the effectiveness and generality of our method, we also performed distillation on the Diffusion Policy\cite{chi2023diffusion}, achieving similarly leading results. This ensures the applicability of our approach to the distillation of various diffusion policies. While previous works have demonstrated the superiority of 3D Diffusion Policy~\cite{ze20243d}, the use of 3D often entails significantly higher computational costs, making it unsuitable for various task scenarios and incapable of meeting the computational demands of lightweight embedded devices such as NVIDIA Jetson. Therefore, we believe that validating our method on Diffusion Policy~\cite{chi2023diffusion} is crucial.

\subsection{Experimental Setup}

\noindent\textbf{Datasets.} We conducted experimental validation on the Robomimic~\cite{mandlekar2021matters} dataset, we selected the task 'Square', which is considered more challenging for evaluation, while excluding the simpler tasks, Lift and Can. 

\noindent\textbf{Baselines.} To align with the experimental setup in the main text and demonstrate the effectiveness of our SDM Policy model in balancing fast inference speed and accurate action quality, we reproduced the original Diffusion Policy~\cite{chi2023diffusion} and the Consistency Policy~\cite{prasad2024consistency}, which adopts consistency distillation methods, as baselines for comparison.

\noindent\textbf{Evaluation metrics.} During the evaluation, we observed some variations in the success rates across different environment initializations. For this experiment, we ran 3 random seeds, specifically seeds 42, 43, and 44, to mitigate the impact of performance fluctuations. We report the average peak success rate across the three random seeds for each method during training.

\subsection{Efficiency and Effectiveness}

To showcase the effectiveness of our method, we thoroughly evaluated it based on four critical metrics: task success rate, learning efficiency, inference time, and action learning. The SDM Policy demonstrated exceptional performance across all these dimensions.

\noindent \textbf{High accuracy.} Table~\ref{table:2d_Sr} presents the evaluation results for the Square task, comparing DDIM, EDM, Consistency Policy, and our SDM Policy. To ensure a fair comparison, DDIM and EDM were trained with the same number of epochs. Leveraging the conclusion about learning efficiency from section~\ref{exp:ee} in the main text, and to reduce dependence on computational resources, we trained our model using only 50 epochs. Our method achieved better experimental results compared to consistency distillation-based methods.

\begin{table}[htbp]
\caption{\textbf{Comparisons on success rate.} We evaluated the task using three random seeds and reported the average success rate (\%) along with the standard deviation. Our SDM Policy outperformed the current state-of-the-art models in single-step inference, achieving better results than consistency distillation methods. This demonstrates the effectiveness of our approach in RGB-based vision-guided diffusion policies and highlights the generalizability of our SDM Policy model.}
\label{table:2d_Sr}
\vspace{-0.05in}
\centering
\resizebox{0.38\textwidth}{!}{%
\begin{tabular}{l|cc|cccc}
\toprule
Policy & Epochs & NFE & square-ph   \\
\midrule

\textbf{DDIM} & 400 & 10 & \dd{88}{6}   \\
\textbf{EDM} & 400 & 19 & \dd{82}{9}    \\
\textbf{Consistency Policy} & 50 & 1 & \dd{64}{10}        \\
\rowcolor{mycolor!10}
\textbf{SDM Policy} & 50 & 1 & \dd{66}{5}      \\
\bottomrule
\end{tabular}
}
\vspace{-1.2em}
\end{table}

\noindent \textbf{Competitive inference speed.} Table~\ref{table:2d_time} presents the evaluation inference time results for the Square task, comparing EDM, Consistency Policy, and our SDM Policy. The results demonstrate that our method achieves a comparable performance to the current SOTA consistency distillation method while achieving over 18 times inference acceleration compared to the original EDM-based results. 

\begin{table}[htbp]
\caption{\textbf{Comparisons on inference speed.} We evaluated the task using three random seeds and reported the average speed (Hz). Our SDM Policy outperformed the current state-of-the-art models in single-step inference, achieving better results than consistency distillation methods. This demonstrates the effectiveness of our approach in RGB-based vision-guided diffusion policies and highlights the generalizability of our SDM Policy model.}
\label{table:2d_time}
\vspace{-0.05in}
\centering
\tiny 
\resizebox{0.38\textwidth}{!}{%
\begin{tabular}{l|cc|cccc}
\toprule
Policy & Epochs & NFE & square-ph   \\
\midrule

\textbf{EDM} & 400 & 19 & 2.75Hz    \\
\textbf{Consistency Policy} & 50 & 1 & 51.77Hz    \\
\rowcolor{mycolor!10}
\textbf{SDM Policy} & 50 & 1 & 49.13Hz   \\
\bottomrule
\end{tabular}
}
\end{table}

\noindent \textbf{Precise and accurate action.} As shown in Table~\ref{table:2d_action}, we also found that the action quality learned by the SDM Policy is precise, and consistent with the current SOTA methods. This is in line with the previously mentioned higher success rate and learning efficiency, further validating the effectiveness of our approach.

\begin{table}[htbp]
\caption{\textbf{Comparisons on action.} We evaluated the task using three random seeds and reported the average action error and standard deviation. Our SDM Policy outperformed the current state-of-the-art models in single-step inference, achieving better results than consistency distillation methods. This demonstrates the effectiveness of our approach in RGB-based vision-guided diffusion policies and highlights the generalizability of our SDM Policy model.}
\label{table:2d_action}
\vspace{-0.05in}
\centering
\tiny 
\resizebox{0.38\textwidth}{!}{%
\begin{tabular}{l|cc|cccc}
\toprule
Policy & Epochs & NFE & square-ph   \\
\midrule
\textbf{DDIM} & 400 & 10 & \dd{0.096}{0.029}    \\
\textbf{EDM} & 400 & 19 & \dd{0.092}{0.009}    \\
\textbf{Consistency Policy} & 50 & 1 & \dd{0.050}{0.001}    \\
\rowcolor{mycolor!10}
\textbf{SDM Policy} & 50 & 1 & \dd{0.050}{0.000}    \\
\bottomrule
\end{tabular}
}
\end{table}

\section{Further Discussion}

This work has demonstrated that enforcing the minimization of the matching loss between two diffusion distributions provides a direct signal for training the generator, significantly improving distillation performance. The 57 experiments listed in the paper have shown that our method achieves a balance between fast inference speed and high-quality actions. Although we briefly discussed follow-up questions in the main text, we provide a more detailed description here.

First, our average task success rate does not exceed that of the teacher model DP3, with only a subset of tasks surpassing it. Overall, our method achieves performance closer to the teacher model compared to consistency distillation methods. From an information-theoretic perspective, the distilled model trained through knowledge distillation incorporates more effective Knowledge Points and can simultaneously learn multiple Knowledge Points. This results in more stable optimization compared to models trained from scratch, where the teacher model learns sequentially. In future work, we will continue to explore this direction.

In addition, based on the data obtained from testing 57 simulation tasks, our SDM Policy achieves an average speed of 61.75Hz, representing a 6x improvement compared to DP3's 10.9Hz. For practical applications, extending our method to dynamic operations and tasks requiring high-frequency control is critical to adapting to more complex and unpredictable environments. For instance, dexterous hands operating on complex objects such as twisting bottle caps or grasping flexible objects require real-time feedback on the force and position of each finger to adjust the grasping policy dynamically. Precision operations such as threading a needle or assembling electronic components demand higher control accuracy, requiring real-time closed-loop perception and control during the inference process. For grippers, tasks involving dynamic scenarios such as grabbing moving targets on a fast conveyor belt necessitate higher inference frequency to maintain success rates. Additionally, during the grasping process, objects may suddenly slip or change position, and fast inference is essential to quickly adjust actions when environmental feedback changes. These discussions underscore the significance of our method's exploration of fast inference and high-quality action learning. In future work, we will further investigate and address these scenarios and discussions.

\end{document}